\documentclass[10pt,journal]{IEEEtran}

\usepackage[final]{changes}
\definecolor{darkgreen}{RGB}{0,111,0}
\usepackage{graphicx}
\usepackage{epstopdf}
\usepackage{multirow}

\ifCLASSOPTIONcompsoc
  \usepackage[nocompress]{cite}
\else
  \usepackage{cite}
\fi

\ifCLASSINFOpdf
\else
\fi

\usepackage[noend]{algorithmic}
\usepackage{algorithm}
\usepackage{draftwatermark}

\begin{document}
\SetWatermarkAngle{0}
\SetWatermarkColor{black}
\SetWatermarkLightness{0.5}
\SetWatermarkFontSize{9pt}
\SetWatermarkVerCenter{20pt}
\SetWatermarkText{\parbox{20cm}{
\centering This is the accepted version of the article Straka, Z.; Svoboda, T. \& Hoffmann, M. (2023), 'PreCNet: Next-frame video prediction based on predictive coding', IEEE Transactions on Neural Networks and Learning Systems. DOI: https://doi.org/10.1109/TNNLS.2023.3240857. \copyright IEEE
}}

\title{PreCNet: Next-Frame Video Prediction Based on Predictive Coding}

\author{Zdenek~Straka,~\IEEEmembership{}
        Tom\'a\v s~Svoboda,~\IEEEmembership{Member,~IEEE,}
        and~Matej~Hoffmann,~\IEEEmembership{Member,~IEEE}
\IEEEcompsocitemizethanks{\IEEEcompsocthanksitem Z. Straka, T. Svoboda, and M. Hoffmann are with the Department of Cybernetics, Faculty of Electrical Engineering, Czech Technical University in Prague, 12135 Prague, Czech Republic.\protect\\
E-mail: {straka.zdenek{/}svobodat{/}matej.hoffmann@fel.cvut.cz}

This is the accepted version of the article Straka, Z.; Svoboda, T. \& Hoffmann, M. (2023), 'PreCNet: Next-frame video prediction based on predictive coding', IEEE Transactions on Neural Networks and Learning Systems. DOI: https://doi.org/10.1109/TNNLS.2023.3240857.

\copyright 2023 IEEE.  Personal use of this material is permitted.  Permission from IEEE must be obtained for all other uses, in any current or future media, including reprinting/republishing this material for advertising or promotional purposes, creating new collective works, for resale or redistribution to servers or lists, or reuse of any copyrighted component of this work in other works.
}
}


\IEEEtitleabstractindextext{
\begin{abstract}
Predictive coding, currently a highly influential theory in neuroscience, has not been widely adopted in machine learning yet. In this work, we transform the seminal model of Rao and Ballard (1999) into a modern deep learning framework while remaining maximally faithful to the original schema. The resulting network we propose (PreCNet) is tested on a widely used next frame video prediction benchmark, which consists of images from an urban environment recorded from a car-mounted camera, and achieves state-of-the-art performance. Performance on all measures (MSE, PSNR, SSIM) was further improved when a larger training set (2M images from BDD100k), pointing to the limitations of the KITTI training set. This work demonstrates that an architecture carefully based in a neuroscience model, without being explicitly tailored to the task at hand, can exhibit exceptional performance. 
\end{abstract}

}

\maketitle

\IEEEdisplaynontitleabstractindextext

\IEEEpeerreviewmaketitle

\ifCLASSOPTIONcompsoc
\IEEEraisesectionheading{\section{Introduction}\label{sec:introduction}}
\else
\section{Introduction}
\label{sec:introduction}
\fi

\IEEEPARstart{P}{redicting} near future is a crucial ability that every agent---human, animal, or robot---needs for survival in a dynamic and complex environment. Just for safely crossing a busy road, one needs to anticipate the future position of cars, pedestrians, as well as consequences of own actions. Machines are still lagging behind in this ability. For deployment in such environments, it is necessary to overcome this gap and develop efficient methods for foreseeing the future.

One candidate approach for predicting near future is predictive coding---a popular theory from neuroscience. The basic idea is that the brain is a predictive machine which anticipates incoming sensory inputs and only the prediction errors---unpredicted components---are used for the update of an internal representation.
In addition, predictive coding tackles another important aspect of perception: how to efficiently encode redundant sensory inputs~\cite{huang2011predictive}. 
Rao and Ballard proposed and implemented a hierarchical architecture~\cite{rao1999predictive}---which we will refer to as \textit{predictive coding schema} (see Section \ref{sec:PC_schema} for details)---that explains certain important properties of the visual cortex: the presence of oriented edge/bar detectors and extra-classical receptive field effects. 
This schema has influenced several works on human perception and neural information processing in the brain (see e.g.,~\cite{spratling2008reconciling,stefanics2014visual,summerfield2009expectation,spratling2010predictive}; for reviews~\cite{friston2018does,huang2011predictive,clark2013whatever}).

In this work, our goal was to remain as faithful as possible to the \textit{predictive coding schema} but cast it into a modern deep learning framework. We thoroughly analyze how the conceptual architecture is preserved. To demonstrate the performance, we chose a widely used benchmark---next frame video prediction---for the following reasons. First, large datasets of unlabeled sequences are available and this task bears direct application potential. Second, this task is an instance of unsupervised representation learning, which is currently actively researched (e.g.,~\cite{mathieu2016deep}). Third, the complexity of the task can be scaled, for example by performing multiple frame prediction (frames are anticipated more steps ahead). 
On a popular next frame video prediction benchmark, our---strongly biologically grounded---network achieves state-of-the-art performance. 
In addition to commonly used training dataset (KITTI), we trained the model on a significantly bigger dataset which improved the performance even further.

We summarize our contributions as follows. First, in this work, the seminal predictive coding model of Rao and Ballard~\cite{rao1999predictive} has been cast into a modern deep learning framework, while remaining as faithful as possible to the original schema. Second, we tested our architecture (PreCNet) on a widely used next frame video prediction benchmark (KITTI with 41k images for training, Caltech Pedestrian Dataset for testing) and outperformed most of the state-of-the-art methods and achieved 2nd-3rd rank when measured with the Structural Similarity Index (SSIM)---a performance measure that should best correlate with human perception.
Third, performance on all three measures (MSE, PSNR, SSIM) was significantly improved when a larger training set (BDD100k with 2M images) than usually used by the community (KITTI) was employed.

This article is structured as follows. The Related Work section overviews models inspired by predictive coding and state-of-the-art methods for video prediction. This is followed by the Architecture section where we describe our model and compare it in detail with the original Rao and Ballard schema~\cite{rao1999predictive} and PredNet~\cite{lotter2017deep}---a model for next frame video prediction inspired by predictive coding. In Section~\ref{sec:experiments}, we detail the datasets, performance metrics, and our experiments in next and multiple frame video prediction. This is followed by Conclusion, Discussion, and Future Work.
All code and trained models used in this work are available at \cite{github-precnet}.

\section{Related Work}
This section starts with a summary of predictive coding-inspired machine learning models. This is followed by an overview of state-of-the-art methods for video prediction.  

\subsection{Predictive coding models}
\label{sec:pred_cod_models}
 In this section, we will focus on predictive coding-inspired machine learning models. A reader who is interested in the application in computational and theoretical neuroscience may find useful reviews~\cite{huang2011predictive,clark2013whatever,spratling2017review} and references~\cite{rao1999predictive,stefanics2014visual,summerfield2009expectation,spratling2010predictive,friston2005theory,spratling2008reconciling}.
 Predictive coding, a theory originating in neuroscience, is more a general schema (with certain properties) than a concrete model. Therefore, no ``correct'' model of predictive coding is available to date. 
 In this work, by predictive coding, we will understand a well defined schema proposed by Rao and Ballard~\cite{rao1999predictive}, which was also implemented as a computational model (see Section \ref{sec:PC_schema} for a description of the schema).
This schema, which is highly influential in neuroscience, embodies crucial ideas of the predictive coding theory.

We will relate predictive coding-inspired machine learning models to the schema by Rao and Ballard and analyze which properties of the original are preserved and which are not.
A detailed comparison of our deep neural network---intended to be as faithful as possible to the Rao and Ballard schema---will be presented in a separate Section \ref{sec:PC_vs_RB}. The models with static inputs and sequences will be presented separately.

\subsubsection{Models with static inputs}
Song et al.~\cite{song2018fast} proposed Fast Inference Predictive Coding model (FIPC) model for image representation and classification which extends the schema by Rao and Ballard by (i) a regression procedure with fast inference during testing and (ii) a classification layer which directs representation learning to achieve discriminative features.
An important part of predictive coding theory is the existence of prediction error neurons along with representational neurons (see~\cite{rao1999predictive,clark2013whatever}).
Models~\cite{spratling2017hierarchical,han2018deep,wen2018deep} intended for object recognition in natural images have these two distinct neural populations, however, their training is not based on the prediction error minimization used in predictive coding.
A generative model by Dora et al.~\cite{dora2018deep} for inferring causes underlying visual inputs does not follow the division into the error and representational neurons. However, the model is trained, in accordance with predictive coding, to minimize prediction errors. The same authors contributed to the model which extends the predictive coding approach to inference of latent visuo-tactile representations~\cite{struckmeier2019mupnet}, used for place recognition of a biomimetic robot in a simulated environment.     

\subsubsection{Models with sequences as inputs}
Ahmadi and Tani proposed the predictive-coding-inspired variational recurrent neural network~\cite{ahmadi2019novel} (PV-RNN). The network works in a three stage processing cycle: 
(i) producing prediction,
(ii) backpropagating the prediction errors across the network hierarchy,
(iii) updating the internal states of the network to minimize future prediction errors.
The network was used for synchronous imitation between two robots---joint angles and XYZ coordinates of a hand tip were used---and for extracting latent probabilistic structure from a binary output of a simple probabilistic finite state machine.
Using the same three stage predictive coding processing cycle, Choi and Tani developed a predictive multiple spatio-temporal scales recurrent neural network~\cite{choi2018predictive} (P-MSTRNN) for predicting binary image (36x36 pixels) sequences of human whole-body cyclic movement patterns. They also explored how the inferred internal (latent) states can be used for recognition of the movement patterns.
Chalasani and Principe proposed a hierarchical linear dynamical model for feature extraction~\cite{chalasani2013deep}. The model took inspiration from predictive coding and used higher-level predictions for inference of lower-level predictions.
However, all three models do not use the division into the error and representational neurons and consequently use a different schema than Rao and Ballard~\cite{rao1999predictive}.
  
Lotter et al. proposed a predictive neural network (PredNet)  for next-frame video prediction~\cite{lotter2017deep}. The network follows the division into error and representational neurons, but the processing schema is different to the one proposed by Rao and Ballard~\cite{rao1999predictive} and consequently to our model (see Section~\ref{sec:precnet_vs_prednet} for details). Despite the architectural differences from the schema by Rao and Ballard, the network could mimic certain features of biological neurons and perception~\cite{lotter2020neural}.

\subsection{Video prediction models}
Video prediction is an important task in computer vision with a long history. A sequence of images is given and one or multiple following images are predicted (i.e., next and multiple frame video prediction task respectively). As prediction of the next sensory input is inherent to predictive coding, next frame video prediction provides a natural use case to benchmark the performance of our neural network architecture. Therefore, we will focus predominantly on a brief review of recent work with state-of-the-art performance on next frame video prediction and---wherever feasible---we will quantitatively compare the performance (see Section \ref{sec:quant_analysis}). 

Many of the methods for video prediction produce blurred predictions. As blurriness is undesirable, Matthieu et al.~\cite{mathieu2016deep} proposed a gradient difference loss function which is minimized when the gradient of the actual and predicted image is the same. This loss function was then combined with adversarial learning. 
Byeon et al.~\cite{byeon2018contextvp} showed with their LSTM-based architecture that direct connection of each predicted pixel with the whole available past context led to decreasing prediction uncertainty on pixel level and therefore also reduced blurriness.         
Reda et al.~\cite{reda2018sdc} suggested that blurriness is amplified by using datasets with lack of large motion and small resolution. Therefore, they used video games (GTA-V and Battlefield-1) for generation of a large high-resolution dataset with large enough motion (testing was performed on natural sequences).
The dataset was then used for training of a model which combines a kernel-based approach with usage of optical flow.
In addition to optical flow estimation, a model by Lu et al. \cite{lu2021video} used pixel generation and adversarial training. Gao et al.~\cite{gao2019disentangling} proposed a model which performed generation of the future frames in two steps. Firstly, a flow predictor was used for warping the non-occluded regions. Then, the occluded regions were in-painted by a separate network. 
A method by Liu et al.~\cite{liu2017video} did not use optical flow directly, however, a deep network was trained to synthesize a future frame by flowing pixel values from the given video frames. This self-supervised method was also used for interpolation.
Similarly to Gao et al.~\cite{gao2019disentangling}, Hao et al.~\cite{hao2018controllable} proposed a two-stage architecture. However, the input of a network contained, in addition, sparse motion trajectories (automatically extracted for video prediction). First, the network produced a warped image that respected the given motion trajectories. In the second stage, occluded parts of the image were hallucinated and color change was compensated. 

Villegas et al.~\cite{villegas2017learning} introduced a model which first performed human pose detection and its future evolution. Then, the predicted human poses were used for future frames generation. 

Finn et al.~\cite{finn2016unsupervised} proposed a model that next to visual inputs takes actions of the robot into account. This action-conditioned model learned to anticipate pixel motions relatively to the previous frame.

A Conditionally Reversible Network (CrevNet) proposed by Yu et al.~\cite{Yu2020Efficient} uses a bijective two-way autoencoder, based on convolutional networks, for encoding and decoding input frames. Feature maps obtained from the autoencoder are then used as an input to a ConvRNN-based predictor. The transformed feature maps by the predictor are then decoded by the autoencoder and outputted as predicted frames. Chang et al. proposed Information Preserving Spatiotemporal Predictive Model \cite{chang2021iprnn} (IPRNN) which 
used skip connections from encoders to decoders. As a decoder has access to information from an encoder, the information loss is reduced. The model used stacked spatiotemporal gated recurrent units which took the encoded states as input. Yuan et al. integrated the attention mechanism into a convolutional LSTM network \cite{yuan2021novel}. This model was further extended by pixel restoration of the input images to the predictions and denoted as Deep Pixel Restoration AttConvLSTM (DPRAConvLSTM) model. Attention mechanism was also effectively used for human-skeleton motion prediction \cite{shu2021spatiotemporal,zhang2022skip}. 

Some other state-of-the-art architectures are based on generative adversarial networks (GANs). The GAN by Kwon and Park~\cite{kwon2019predicting} is trained to anticipate both future and past frames. The GAN proposed by Liang et al.~\cite{liang2017dual} is trained to consistently predict future frames and pixel-wise flows using a dual learning mechanism. Vondrick et al.~\cite{vondrick2016generating} proposed GAN for generation of image sequences which unravels foreground from the background of the images. A video prediction network proposed by Jin et al.~\cite{jin2020exploring} integrates generative adversarial learning with usage of spatial and temporal wavelet analysis modules.   

The stochastic nature of natural video sequences makes it impossible to predict the future sequence perfectly. Models such as~\cite{lee2018savp,babaeizadeh2018stochastic,denton2018stochastic} attempt to deal with that by generating multiple possible futures. The objective is to predict frame sequences which are: (i) diverse, (ii) perceptually realistic, and (iii) a plausible continuation of the given input sequence or  image~\cite{lee2018savp}.
Therefore, this task is different from deterministic video frame prediction whereby the model is intended to produce only a single frame or sequence best fitting the actual future.

Some of the mentioned works~\cite{liang2017dual,vondrick2016generating,lotter2017deep,Yu2020Efficient} also demonstrated that the representations which were learned during next frames video prediction training could be used for supervised learning tasks (e.g., human action recognition). 
\section{Architecture}
This section starts with a description of the \textit{predictive coding schema} which was proposed by Rao and Ballard~\cite{rao1999predictive}. This is followed by a detailed description of our model. The section is closed by a comparison of our model with related models: (i) a hierarchical network for predictive coding proposed by Rao and Ballard, (ii) PredNet -- a deep network for next frame video prediction inspired by predictive coding.

\subsection{Predictive coding schema}
\label{sec:PC_schema}
\begin{figure*}[!ht]
\centering
\includegraphics[width=5in]{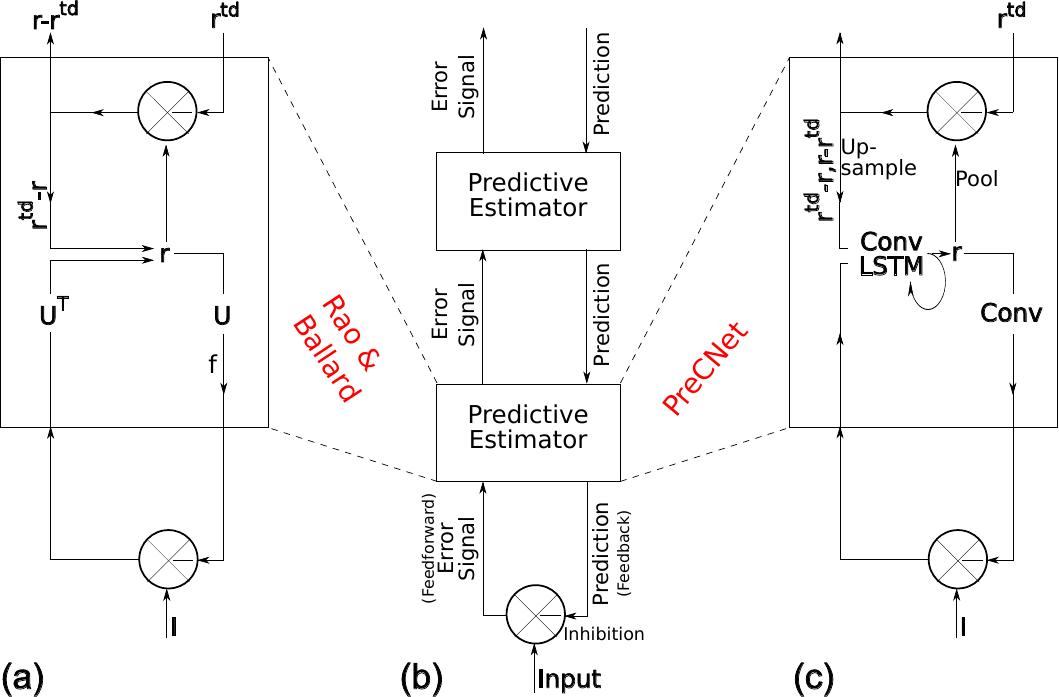}
\caption{\textbf{Comparison of the hierarchical network for predictive coding by Rao and Ballard and our PreCNet.} \textbf{(a)} \textbf{Components of a Predictive Estimator (PE) module of the model by Rao and Ballard}, composed of feedforward neurons encoding the synaptic weights $\mathbf{U}^T$ , neurons whose responses $\mathbf{r}$ maintain the current estimate of the input signal, feedback neurons encoding $\mathbf{U}$ and conveying the prediction $f(\mathbf{U}\mathbf{r})$ to the lower level, and error-detecting neurons computing the difference $(\mathbf{r} - \mathbf{r}^{td})$ between the current estimate $\mathbf{r}$ and its top-down prediction $\mathbf{r}^{td}$ from a higher level. \textbf{(b)} \textbf{General architecture of the hierarchical predictive coding model.} At each hierarchical level, feedback pathways carry predictions of neural activity at the lower level, whereas feedforward pathways carry residual errors between the predictions and actual neural activity. These errors are used by the PE at each level
to correct its current estimate of the input signal and generate the next prediction.  \textbf{(c)} \textbf{Components of a PE module of PreCNet architecture} (see Section \ref{sec:PC_vs_RB}). Figures \textbf{(a)} and \textbf{(b)} redrawn from~\cite{rao1999predictive}, their captions with minor modification from~\cite{rao1999predictive}.}
\label{fig:compar_RB_vs_PreCNet}
\end{figure*}

Motivated by crucial properties of the visual cortex, Rao and Ballard have proposed a hierarchical \textit{predictive coding schema} with its implementation~\cite{rao1999predictive}.   
According to this schema, throughout the hierarchy of visual processing, ``feedback connections from a higher- to a lower-order visual cortical area carry predictions of lower-level neural activities, whereas the feedforward
connections carry the residual errors between the predictions and the actual lower-level activities''\cite{rao1999predictive}. The residual errors are used to reduce the prediction error in the following moment (see Fig.~\ref{fig:compar_RB_vs_PreCNet}, \textbf{(b)}).

This schema was directly turned into a computational model in~\cite{rao1999predictive} (see Fig.~\ref{fig:compar_RB_vs_PreCNet}, \textbf{(a)}). The feedback connection from higher-level to lower-level Predictive Estimator (PE) carries the top-down prediction $\mathbf{r}^{td}$ of the lower-level PE activity $\mathbf{r}$. The residual error $\mathbf{r}-\mathbf{r}^{td}$ is sent back via feedforward connections to the higher-level PE. The same error with opposite sign, $\mathbf{r}^{td}-\mathbf{r}$, affects the following PE activity $\mathbf{r}$ (see  Fig.~\ref{fig:compar_RB_vs_PreCNet}, \textbf{(a)}). The bottom-level PE produces a prediction of the visual input.

Drawing on the \textit{predictive coding schema}, we propose the \textit{Predictive Coding Network (PreCNet)} (see Fig.~\ref{fig:compar_RB_vs_PreCNet}, \textbf{(c)}). In contrast with the model by Rao and Ballard (compare parts \textbf{(a), (c)} of Fig.~\ref{fig:compar_RB_vs_PreCNet}), PreCNet uses a modern deep learning framework (see Section \ref{sec:model_descr} for details of PreCNet architecture and Section \ref{sec:PC_vs_RB} for a more detailed comparison of both models). This has enabled us to create a model based on the \textit{predictive coding schema} with state-of-the-art performance, as demonstrated on the next-frame video prediction benchmark.   

\subsection{Description of PreCNet model (ours)} \label{sec:model_descr}
The structure, computation of prediction and states, and training of the model is detailed below. 
\subsubsection{Structure of the model}
\begin{figure}[!t]
\centering
\includegraphics[width=0.45\textwidth]{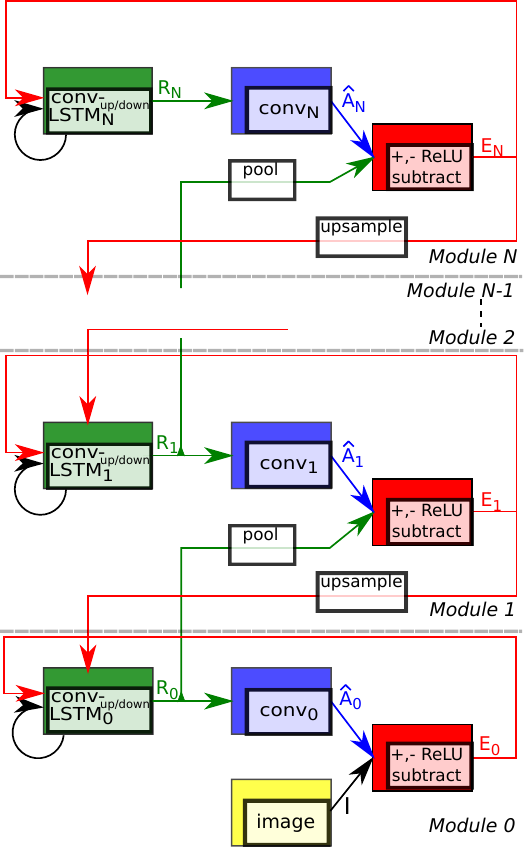}
\caption{\textbf{Modular architecture of PreCNet.} The highest module misses connections upwards. Main parts of each module are a representation layer (green), decoding layer (blue) and error representations (red). See the text and Alg.\ref{alg:calc_activ} for more details.}
\label{fig:precnet_arch}
\end{figure}

The model, shown in Fig.~\ref{fig:precnet_arch}, consists of $N+1$ hierarchically organized modules\footnote{The model is the same as in Fig.~\ref{fig:compar_RB_vs_PreCNet}, \textbf{(c)}. However, in order to enable direct comparison with Rao and Ballard model, it was redrawn in a different arrangement for Fig.~\ref{fig:compar_RB_vs_PreCNet}, \textbf{(b)}. The PE from the model of~\cite{rao1999predictive} is not equivalent to the ``Module'' in Fig.~\ref{fig:precnet_arch}.  See Fig.~\ref{fig:compar_PreCNet_vs_PredNet}, \textbf{(b)}, \textbf{(e)} for a comparison.}. A module $i\in\{0,1,\dots, N\}$ consists of the following components:
\begin{itemize}
    \item A \textbf{representation layer} is a convolutional LSTM ($\textrm{convLSTM}_i$) layer (see~\cite{xingjian2015convolutional,hochreiter1997long}) with output state $R_i$ (alternatively\footnote{For representation layer states we used both small $\mathbf{r}$ and capital letter $R$. Small $\mathbf{r}$ corresponds to formalism from~\cite{rao1999predictive}, capital $R$ was used in~\cite{lotter2017deep}.} $\mathbf{r}$). The convLSTM followed dynamics from commonly used ``No peepholes'' LSTM variant \cite{greff2016lstm} (see \textit{Supplementary materials -- convLSTM} for details). Technically, it consists of two convolutional LSTM layers ($\textrm{convLSTM}_i^{up/down}$) which share hidden and cell states ($R_i, C_i$) but differ in the input ($E_i$ vs. $E_{i+1}$). The input, forget, and output gates use hard sigmoid as an activation function. During calculation of the final (hidden) and cell states, hyperbolic tangent is used.  
    
    \item An  \textbf{error representation} consists of the Rectified Linear Units (ReLU) whose input is obtained by merging errors $\mathit{PREDICTION} - \mathit{ACTUAL} \ \mathit{STATE}$  and $\mathit{ACTUAL} \ \mathit{STATE}  - \mathit{PREDICTION}$.
    The state of the error representation is denoted as $E_i$. 
    
    \item A \textbf{decoding layer} is a convolutional ($\textrm{conv}_i$) layer with output state $\hat{A}_i$. It uses ReLU as an activation function.
    \item An \textbf{upsample layer}, which uses nearest-neighbor method, upscales its input by factor 2. This layer is not present in the module 0.
    \item A \textbf{max-pooling layer} which downscales its input by a factor 2. This layer is not present in the module 0.
\end{itemize}
  
\subsubsection{Computation of the prediction and states}
\label{sec:comp_pred_states}
In every time step, PreCNet outputs a prediction of the incoming image. The error of the prediction is then used for the update of the states (see also Fig.~\ref{fig:next_frame_pred}). The computation in every time step can be divided into two phases:
\begin{enumerate}
    \item \textbf{Prediction phase}. The information flow goes iteratively from a higher to a lower module. At the end of this phase (at Module 0), the prediction of the incoming input image $\hat{A}_0$ is outputted.     
    \item \textbf{Correction phase}. In this phase, the information flow goes iteratively up. The error between the prediction and actual input is propagated upward. 
\end{enumerate}
In a nutshell, a representational layer (with state $R_i$) represents a prediction of the image $I$ ($i=0$) or a pooled convLSTM state $R_{i-1}$ from the module bellow ($i>0$). The decoding layer transforms the representation $R_i$ into the prediction $\hat{A}_i$. The error representation units $E_i$ then depend on the error of the prediction $\hat{A}_i$ (difference between the prediction $\hat{A}_i$ and the image $I$ or the pooled state $R_{i-1}$). The computation is completely described in Alg. \ref{alg:calc_activ}.

\begin{algorithm}
\caption{Calculate PreCNet states at time $t$, assume $N>0$. Merging of states A and B is denoted by putting them into curly brackets \{A, B\}. See \textit{Supplementary materials -- convLSTM} for a detailed convLSTM description.}
\label{alg:calc_activ}
\begin{algorithmic}
\REQUIRE Image $I^t$, previous ($t-1$) hidden and cell states $R^{t-1}_l, C^{t-1}_l$ of the representation layers $l \in \{0,1,\dots,N\}$, previous error state $E^{t-1}_{N}$ of the (top) module $N$, maximum pixel value $\mathit{pix}^{max}$. 
For $t=1$, the previous states $R_l^{0}, C_l^{0}, E_N^{0}$ are initialized to zero.

\ \FOR{$l = N, N-1, \dots, 0$ \COMMENT{Iterate top-down through the modules}} 
\IF{$l == N$ \COMMENT{Update the states in the top module}}
\STATE  $R^{t}_l, C^{t}_l \leftarrow \textrm{convLSTM}_l^{down}(R^{t-1}_{l}, C^{t-1}_{l}, E^{t-1}_{l})$
\STATE $\hat{A}^t_l \leftarrow \textrm{conv}_l(R^{t}_l)$
\STATE $E^{t}_l \leftarrow \textrm{ReLU}(\{\hat{A}^t_l -  \textrm{pool}(R^{t-1}_{l-1}),\textrm{pool}(R^{t-1}_{l-1}) - \hat{A}^t_l\})$
\ENDIF
\STATE

\IF{$l \neq N$ \AND $l \neq 0$ \COMMENT{Update the states in the ``middle'' module $l$}}
\STATE  $R^{t}_l, C^{t}_l \leftarrow \textrm{convLSTM}_l^{down}(R^{t-1}_{l}, C^{t-1}_{l}, \textrm{upsample}(E^{t}_{l+1}))$
\STATE $\hat{A}^t_l \leftarrow \textrm{conv}_l(R^{t}_l)$
\STATE $E^{t}_l \leftarrow \textrm{ReLU}(\{\hat{A}^t_l -  \textrm{pool}(R^{t-1}_{l-1}),\textrm{pool}(R^{t-1}_{l-1}) - \hat{A}^t_l\})$
\ENDIF
\STATE

\IF{$l==0$ \COMMENT{Update the states in the bottom module}}
\STATE  $R^{t}_l, C^{t}_l \leftarrow \textrm{convLSTM}_l^{down}(R^{t-1}_{l}, C^{t-1}_{l}, \textrm{upsample}(E^{t}_{l+1}))$
\STATE $\hat{A}^t_l \leftarrow \textrm{min}(\textrm{conv}_l(R^{t}_l),\mathit{pix}^{max})$
\STATE $E^{t}_l \leftarrow \textrm{ReLU}(\{\hat{A}^t_l -  I^t,I^t - \hat{A}^t_l\})$
\ENDIF

\ENDFOR
\STATE
\FOR{$l = 0,1, \dots, N$ \COMMENT{Iterate bottom-up through the modules}}
\IF{$l ==  0$}
\STATE  $R^{t}_l, C^{t}_l \leftarrow \textrm{convLSTM}_l^{up}(R^{t}_{l}, C^{t}_{l}, E^{t}_{l})$
\ENDIF
\STATE

\IF{$l \neq  0$ \AND $l \neq N$}
\STATE $E^{t}_l \leftarrow \textrm{ReLU}(\{\hat{A}^t_l -  \textrm{pool}(R^{t}_{l-1}),\textrm{pool}(R^{t}_{l-1}) - \hat{A}^t_l\})$
\STATE  $R^{t}_l, C^{t}_l \leftarrow \textrm{convLSTM}_l^{up}(R^{t}_{l}, C^{t}_{l}, E^{t}_{l})$
\ENDIF
\STATE

\IF{$l ==  N$}
\STATE $E^{t}_l \leftarrow \textrm{ReLU}(\{\hat{A}^t_l -  \textrm{pool}(R^{t}_{l-1}),\textrm{pool}(R^{t}_{l-1}) - \hat{A}^t_l\})$
\ENDIF
\ENDFOR

\end{algorithmic}
\end{algorithm}

\subsubsection{Training of the model}
The model is trained by minimizing weighted prediction errors through the time and hierarchy~\cite{lotter2017deep}. The loss function is defined as 
\begin{equation}
\label{eq:Ltrain}
L_{\mathit{train}}=\sum_{m=1}^{M}L_{\mathit{seq}}(m),    
\end{equation}

\begin{equation}
\label{eq:Lseq}
L_{\mathit{seq}} (j)=\sum^{l_s}_{t=1} \mu_t \sum^N_{l=0} \frac{\lambda_l}{n_l} \sum_{i=1}^{n_l} E^t_l(i),    
\end{equation}
where $L_{seq}(m)$ is loss of the $m^{th}$ sequence, $E_l^t(i)$ is the error of the $i^{th}$ unit in the module $l$ at time $t$, $M$ is a number of image sequences, $l_s$ is a length of a sequence, $N+1$ is a number of modules, $\mu_t$, $\lambda_l$ are time and module weighting factors, $n_l$ is the number of error units in the $l^{th}$ module.
The mini-batch gradient descent was used for the minimization.

\subsection{Comparison of PreCNet with other models}
\label{sec:comparison_PreCNet}
We will compare our model with the \textit{predictive coding schema}~\cite{rao1999predictive}, the Fast Inference Predictive Coding model (FIPC)~\cite{song2018fast}, and PredNet~\cite{lotter2017deep}.

\subsubsection{Comparison of PreCNet and Rao and Ballard model}\label{sec:PC_vs_RB}
PreCNet uses the same schema as the model by Rao and Ballard (see Fig.~\ref{fig:compar_RB_vs_PreCNet} and Section \ref{sec:PC_schema}). However, as PreCNet is couched in a modern deep learning framework and uses video sequences as inputs, there are inevitably some differences. The crucial differences are:
\begin{itemize}
    \item Dynamic vs. static inputs. In contrast with PreCNet and image sequences as inputs, the model by Rao and Ballard takes static images as inputs. An extension to next frame video prediction should be possible~\cite{rao1997dynamic}\footnote{By using recurrent transformation of the representation layer states $\hat{\mathbf{r}}(t+1)=f(V\mathbf{r}(t))$, where  $\hat{\mathbf{r}}(t+1)$ is the prediction of the next state $\mathbf{r}(t+1)$ made at time $t$, $f$ is a nonlinear function, and $V$ are synaptic recurrent weights.},
    but has not been completely demonstrated (in~\cite{rao1999optimal}, a model with only one level of hierarchy is employed).
    These recurrent connections resemble the recurrent connections inside PreCNet representation (convLSTM) layer.
    
    \item Different building blocks. 
    PreCNet, in contrast to Rao and Ballard model, uses modern deep learning blocks -- convolutional and convLSTM layers. In addition, the error representation of PreCNet consists of merged positive, $\mathit{PREDICTION} - \mathit{ACTUAL} \ \mathit{STATE}$, and negative, $\mathit{ACTUAL} \ \mathit{STATE} - \mathit{PREDICTION}$, error populations~\cite{lotter2017deep}. 
    These two populations are also used in the model of Rao and Ballard, however, they are not merged and are used separately. In contrast to PreCNet, ReLU is not applied to the error populations. 
    
    \item Different update of representation states. Representation layer states of the Rao and Ballard model are determined by a first-order differential equation. The states are updated until they converge. In contrast, PreCNet's representation layer states are calculated by convLSTM. 
    
    To update the representation states $\mathbf{r}$ of the model by Rao and Ballard, the ``bottom-up'' difference between the prediction of the PE and the actual input ($I-f(U\mathbf{r})$ in Fig.~\ref{fig:compar_RB_vs_PreCNet}, \textbf{(a)}) followed by fully connected layer and the ``top-down'' difference between the predicted state by the higher PE and the actual state of the PE  ($\mathbf{r}^{td}-\mathbf{r}$ in Fig.~\ref{fig:compar_RB_vs_PreCNet}, \textbf{(a)}) are used simultaneously. 
    PreCNet also uses both differences for computation of the new representation states $\mathbf{r}$, however, not simultaneously; one difference is used by the $\textrm{convLSTM}^{down}$ during the prediction phase, the second is used by the $\textrm{convLSTM}^{up}$ during the correction phase (notice that the $\textrm{convLSTM}^{down}$ and $\textrm{convLSTM}^{up}$ share cell and hidden unit states).
    
    \item One vs. mutliple PEs on one level. There are multiple PEs in one level of the model by Rao and Ballard. Higher level PEs progressively operate on bigger spatial areas than the lower level PEs. PreCNet has one PE in each level of the hierarchy. 
    
    \item Intensity of interaction between the Predictive Estimators (PEs).
     Each PE of PreCNet is updated just two times during one time step (one input image). Once during the Prediction (top-down) phase and once during the Correction (bottom-up) phase. This means that each PE interact with its neighbour just two times during one time step. On the contrary, the PEs of the model by Rao and Ballard interact with each other many times (until their representation states converge) during one time step.
     As PreCNet uses the deep learning approach, which is more computationally demanding, such intensive interaction between the PEs is not possible. 
    
    \item Minimizing error in all levels vs. only the bottom level error. Errors in all levels of the model by Rao and Ballard are minimized. However, PreCNet has achieved better results when only the bottom level error---the difference between the predicted and the actual image---was minimized (see the setting of parameter $\lambda_i$ in Section \ref{sec:netw_param}).  
\end{itemize}

\subsubsection{Comparison of PreCNet and FIPC}
As FIPC~\cite{song2018fast} is mainly an extension of the original Rao and Ballard model by a procedure for regression mapping with fast inference at test time, it shares many properties with the Rao and Ballard model. The most important differences between PreCNet and FIPC are: 
\begin{itemize}
    \item Different building blocks. Main building blocks of PreCNet are convolutional and convolutional LSTM networks. FIPC main basic building blocks, similarly to Rao and Ballard model, are simple feedforward networks. This might be limiting for usage on large-scale images and video sequences.
    \item Dynamic vs. static inputs. PreCNet takes image sequences as inputs and predicts next frames. FIPC, identically to Rao and Ballard model, works with static images and is trained for their classification and feature representation.
    \item Fast inference at test time. During testing, the trained network works as a feedforward network (a subset of weights is used) with class labels as outputs. Therefore, during test time the network does not follow the \textit{predictive coding schema}.
    \item Intensity of interaction between the Predictive Estimators (PEs) during training. FIPC Predictive Estimators, similarly to Rao and Ballard model, interact with each other many times during one training time step. For PreCNet, it is only two times (see Section~\ref{sec:PC_vs_RB}). 
    \item Classification layer. FIPC added to the \textit{predictive coding schema} a classification layer which helps to learn more discriminative features for a given classification task. On the other hand, PreCNet is completely self-supervised.
\end{itemize}

\subsubsection{Comparison of PreCNet and PredNet}\label{sec:precnet_vs_prednet}
PredNet, a state-of-the-art deep network for next frame video prediction~\cite{lotter2017deep}, is also inspired by the model by Rao and Ballard. PredNet and PreCNet (which we propose) are similar in these aspects:
\begin{itemize}
    \item Building blocks: error representations, convolutional, and convolutional LSTM networks.
    \item Training procedure. For the next frame video prediction task, most training parameters, such as input sequence length and batch size, of PreCNet are taken from PredNet\footnote{The motivation was two-fold. Firstly, we wanted to make it clear that the significant improvement of PreCNet over PredNet is not caused by better choice of training parameters. Secondly, few trials with other parameter values that we tried did not lead to significantly better results.}.
    \item Number of trainable parameters. For training on the KITTI dataset, PreCNet had approximately 7.6M and PredNet 6.9M trainable parameters. We tested also PreCNet-small with 0.8M parameters (see Table~\ref{tab:kitti_performance} for results).
\end{itemize}

\begin{figure*}[!t]
\centering
\includegraphics[width=7in]{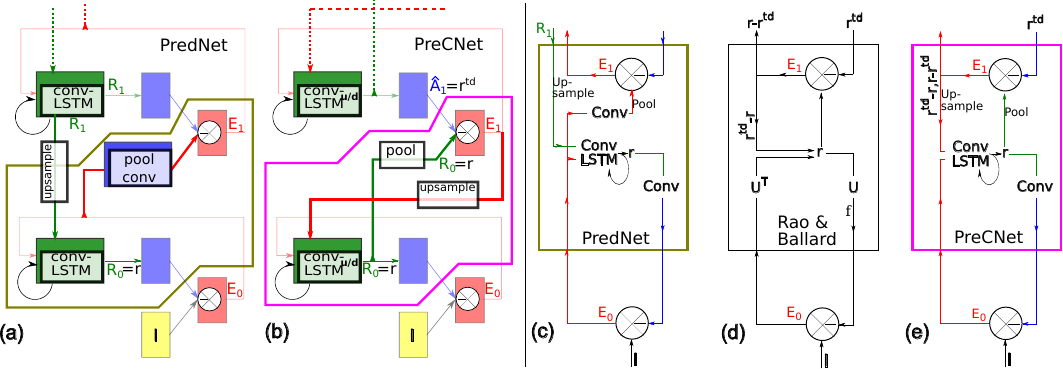}
\caption{\textbf{Comparison of PredNet and PreCNet.} In \textbf{(a), (b)}, the differences (connections between the blocks, some building blocks) are highlighted.  In \textbf{(c), (d), (e)}, there is a comparison of the Predictive Estimators (PEs) of PredNet, PrecNet and the model by Rao and Ballard. Notice that the input from above in \textbf{(d), (e)}  -- prediction $\mathbf{r}^{td}$ of $\mathbf{r}$ -- is compared with the representation state $\mathbf{r}$ and the error is used for update of the $\mathbf{r}$. The corresponding upper input (blue) of the PredNet is a different entity; it is not related to $\mathbf{r}$ and is also compared with a different entity ($\textrm{Conv}(E_0)$). There is also one more input from above -- representation layer state from above -- which goes directly into $\textrm{ConvLSTM}$ block of PredNet. PreCNet (see \textbf{(e)}) has overcome these differences and follows the same predictive coding scheme as the model by Rao and Ballard.           
Notice the correspondence of the olive, purple polygons (\textbf{(a), (b)}) and the PEs of PredNet and PreCNet (the rectangles in \textbf{(c), (e)}). In order to enhance comprehensibility, some of the labels from \textbf{(a), (b)} were added to \textbf{(c), (d), (e)} and v.v.
See \textit{Supplementary materials -- Schema transformation} to check the correspondence between both (\textbf{(a), (b)} and \textbf{(c), (e)}) ways of visualization.
}   
\label{fig:compar_PreCNet_vs_PredNet}
\end{figure*}

However, there are two crucial properties in which PredNet departs from the \textit{predictive coding schema} (see Fig.~\ref{fig:compar_PreCNet_vs_PredNet}, \textbf{(a), (c), (d)}):
\begin{itemize}
    \item According to the \textit{predictive coding schema}, except for the bottom Predictive Estimator (PE), each PE outputs a prediction of the next lower level PE activity $\mathbf{r}$ (representation layer state). See Section~\ref{sec:PC_schema}.    
    \item No direct connection between two neighboring PE activities $\mathbf{r}_i$ and $\mathbf{r}_{i-1}$ (representation layer states $\mathbf{R}_i$ and $\mathbf{R}_{i-1}$ in formalism of~\cite{lotter2017deep}).  
\end{itemize}

Instead, to remain faithful to the \textit{predictive coding schema}, the building blocks of PreCNet were connected in a significantly different way (see Fig.~\ref{fig:compar_PreCNet_vs_PredNet} for comparison).
These modifications have led to considerably better performance of PreCNet in comparison with PredNet (see Section \ref{sec:quant_analysis}).

\section{Experiments}
\label{sec:experiments}
In this section, the datasets and performance measures are introduced, followed by experiments on next frame and multiple frame video prediction.
Trained models and code needed for replication of all the results presented in the paper (dataset preprocessing, model training and evaluation) are available on a \textbf{GitHub repository}~\cite{github-precnet}.

\subsection{Datasets}
\label{sec:datasets}
All datasets used are visual sequences obtained from a car mounted camera. These scenes include fast movements of complex objects (e.g. cars, pedestrians), new objects coming unexpectedly to the scene, as well as movement of the urban background.

For training, we used two different datasets; KITTI~\cite{Geiger2013IJRR} and BDD100K~\cite{yu2019bdd100k}. For evaluation, we used Caltech Pedestrian Dataset~\cite{CVPR09peds, Dollar2012PAMI}, employing {P}iotr's {C}omputer {V}ision {M}atlab {T}oolbox~\cite{PMT} during preprocessing.
Using of Caltech Pedestrian Dataset for establishing performance enables direct comparison of the models from both training variants.

\begin{itemize}
    \item \textbf{KITTI dataset and its preprocessing:} We followed the preprocessing procedure from~\cite{lotter2017deep}. The frames were center-cropped and resized with bicubic method\footnote{We do not know which resizing method was originally used by Lotter et al.~\cite{lotter2017deep}.} to 128 by 160 pixels size (see the repository for code). We also followed the division categories ``city'', ``residential'' and ``road'' of the KITTI dataset to training (57 recording sessions, approx. 41K of frames) and validation parts in the same way as in~\cite{lotter2017deep}. The dataset has 10 fps frame rate. 
    \item \textbf{Caltech Pedestrian Dataset  and its preprocessing:} Frames were preprocessed in the same way as the frames of KITTI dataset (see above). Videos were downsampled from 30 fps to 10 fps (every 3rd frame was taken). As this dataset was used only for evaluation of the performance, only testing parts (set06-set10) were used (approx. 41K of frames).
    \item \textbf{BDD100K and its preprocessing:} The preprocessing of the dataset was analogous to the preprocessing of Caltech Pedestrian Dataset, including reducing frame rate from 30 to 10 fps. As the size of the whole dataset is very large (roughly 40M frames if 10 fps is used), we had to randomly choose training and validation subsets of the dataset---see the repository for details and chosen videos. We created two variants of the training dataset; a big one with roughly 2M frames (5000 recording sessions) and a small one with similar size like KITTI training dataset (approx. 41K frames, 105 recording sessions). As a validation dataset, we randomly selected a subset of the validation part of BDD100K with approx. 9K frames.  
\end{itemize}

\subsection{Performance measures}
\label{sec:measures}
 For comparison of a predicted with the actual frame, we use standard measures: Mean Square Error (MSE), Peak Signal-to-Noise Ratio (PSNR), and Structural Similarity Index (SSIM)~\cite{wang2004image}. MSE is a simple measure whose low values indicate high similarity between frames. PSNR is a related measure to MSE whose value is desired to be as high as possible. Significant limitation of these two is that their evaluation of similarity between two images does not correlate very well with human judgment (e.g.,~\cite{wang2002image,winkler1999perceptual}). SSIM was created to be more correlated with human perception. SSIM values are bounded to $[-1,1]$ and higher value signifies higher similarity. 

\subsection{Next frame video prediction}
\label{sec:next_frame_pred}
Firstly, the settings of experiments and parameters will be described. This is followed by Quantitative results and Qualitative analysis. Results, achieved by PreCNet, presented in this subsection can be generated by publicly available code~\cite{github-precnet}. Summary and details of the network parameters and training are in \textit{Supplementary materials -- 1 Network and training parameters summary and details}.

\subsubsection{Experimental settings}
\label{sec:exp_variants}
We performed experiments with two settings. In both, the performance of trained models was measured using \textit{Caltech Pedestrian Dataset} (see Section \ref{sec:datasets}) which is commonly used for evaluating next frame video prediction task. This also enabled direct comparison of training on both datasets.  
The training was done on:
\begin{itemize}
    \item \textbf{KITTI dataset.} This setting (i.e., KITTI for training, Caltech Pedestrian Dataset for evaluation) is popular for evaluation of next frame video prediction task and enables good comparison with other state of the art methods.
    \item \textbf{BDD100K dataset.} Randomly chosen subset of the dataset (approx. 2M of frames) was used. The training dataset is significantly larger than KITTI dataset which enables to avoid overfitting.
    We also performed training on smaller BDD100K subset with roughly same size as KITTI training dataset.
\end{itemize}

\subsubsection{Network parameters}
\label{sec:netw_param}
Main parameters of the network are summarized in Table~\ref{tab:netw_params}. In the Table, the parameters of each module in the hierarchy are described in a row. Module weights are in the second column. The following columns contain the number of channels \#chan. (layer size) and filter sizes of decoding (conv) and representation (convLSTM) layers. For a detailed explanation see Section \ref{sec:model_descr}.

For choosing a suitable number of hierarchical modules, layer sizes (number of channels), and module weight factors ($\lambda_i, i \in \{0,1,2\}$), KITTI dataset was used for training. 
We performed a manual heuristic parameter search to minimize mean absolute error (between the predicted and actual frames) on validation set\footnote{If $\lambda_0=1, \lambda_{1,2}=0$ then the mean absolute error between the predicted and actual frames corresponds to  2*loss value (\ref{eq:Lseq}). This is a consequence of division of error representation to negative and positive parts and using of ReLU. For non zero $\lambda_{1,2}$, this does not hold.}. Padding was used to preserve the size in all convolutional layers (including convLSTM). Values of the pixels of the input frames were divided by 255 to make them in the range $[0,1]$. The filter sizes were taken from~\cite{lotter2017deep} (for explanation of this choice, see Section~\ref{sec:train_params}).

\begin{table}[!t]
\caption{\textbf{Network parameters summary.} See text for a description.}
\label{tab:netw_params}
\centering
\begin{tabular}{c||c|c|c|c|c}
& &\multicolumn{2}{c}{$conv_i$} & \multicolumn{2}{c}{$convLSTM_i^{up/down}$}\\
\hline
module & weight $\lambda_i$ & \#chan. & filter size  & \#chan. & filter size \\
\hline
\hline
i=0  & 1 & 3 & 3 & 60 & 3  \\
\hline
i=1  & 0 & 60 & 3 & 120 & 3  \\
\hline
i=2 & 0 &120 & 3 & 240 & 3   \\
\hline
\end{tabular}
\end{table}

To better understand how the number of trainable parameters affects the performance and better comparison with PredNet, we proposed the same model but changed the number of channels in the modules from 60, 120, 240 to 20, 40, and 80 respectively (\textit{PreCNet-small}). The number of parameters was reduced from 7.6M to 0.8M. 
Moreover, we simplified the architecture by (i) replacing all pairs of convolutional LSTMs with shared hidden and cell states---$\textrm{convLSTM}^{up}_i$, $\textrm{convLSTM}^{down}_i$---  by single convolutional LSTMs -- $\textrm{convLSTM}_i$ (\textit{PreCNet-single-LSTMs}), and (ii) by simplifying error blocks \textit{ReLU(\{PREDICTION-ACTUAL, ACTUAL-PREDICTION\})} (see Alg. \ref{alg:calc_activ}) to residual errors \textit{}{PREDICTION-ACTUAL} only (\textit{PreCNet-residual-error}). It was also necessary to modify the sequence loss (\ref{eq:Lseq}) by putting the error values $E^t_l(i)$ into absolute value. 

\subsubsection{Training parameters}
\label{sec:train_params}
Except for training length and learning rate, all the values of the training parameters were same as in~\cite{lotter2017deep} (see Section~\ref{sec:precnet_vs_prednet} for the explanation). The network was trained on input sequences with length $l_s=10$. In the sequences used for training and validation, a frame was generally present in more sequences, meaning that the sequences overlap. During learning, the error related to the first predicted input is ignored ($\mu_{t=1}=0$), since the first prediction is produced before seeing any input frame. Prediction errors related to the following time steps are equally weighted ($\mu_t=\frac{1}{l_s-1}, \textrm{for } t \in \{2,..,l_s\}$).

In each epoch, 500 sequences from the training set were randomly selected to form batches of size 4 and used for weight updates. For validation, 100 randomly selected sequences from the validation set were used in each epoch. 
We used Adam~\cite{adam2015} as an optimization method for stochastic gradient descent on the training loss (\ref{eq:Ltrain}). The values of the Adam parameters $\beta_1,\beta_2$ were set to their default values ($\beta_1=0.9$, $\beta_2=0.999$).

Training parameters for training on both datasets were very similar except for number of training epochs and learning rate setting. For the KITTI and BDD100K training, the learning consists of 1000 and 10000 epochs, respectively. Learning rate was set to $0.001$ and $0.0005$ for first 900, 9900 epochs, respectively\footnote{Learning rate setting $0.001$ for BDD100K training led in two of four cases to rapid increase of training loss in later stages of training. Therefore, the learning rate was changed to $0.0005$.}. Then it was decreased to $0.0001$ for last $100$ epochs. As the BDD100K training set is significantly larger than KITTI training set, the training was longer for BDD100K. The choice of the length of the training and learning rate was based on evolution of validation loss and limited computational resources. It means that validation loss still slightly decreased at the final epochs, however, the benefit was not so significant to continue training and use (limited) computational resources.

\subsubsection{Quantitative results}
\label{sec:quant_analysis}

\begin{figure}[!ht]
\centering
\includegraphics[width=3.49in]{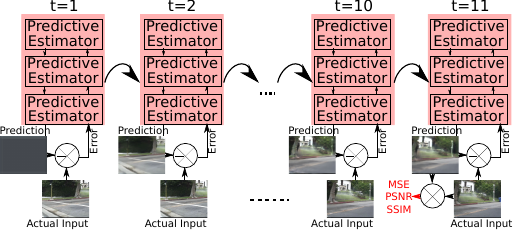}
\caption{\textbf{Next frame video prediction evaluation schema.} In each time step PreCNet outputs next frame prediction. The predicted error is used for update of the network states. After inputting 10 frames (time step $t=11$), the predicted frame is compared---using MSE, PSNR, SSIM---with the actual input. This schema was used for quantitative and qualitative analysis of Next frame video prediction (see Section \ref{sec:next_frame_pred}).}  
\label{fig:next_frame_pred}
\end{figure}

For a quantitative analysis of the performance of the model, we used a standard procedure and measures for evaluating the next frame video prediction. The network obtained a sequence (from Caltech Pedestrian Dataset) of length 10 and then predicted the next frame (see Fig.~\ref{fig:next_frame_pred} for details). Contrary to training and validating sequences, there was no overlap between the two testing sequences of length 11.
This frame is compared to the actual frame using MSE, PSNR and SSIM (see Section \ref{sec:measures}). The overall value of each measure is then obtained as a mean of the calculated values for each predicted frame.   

We performed 10 training repetitions on \textbf{KITTI dataset} (see Section \ref{sec:exp_variants}). The results are summarized in Table~\ref{tab:kitti_repet}. The results show that the learning is stable. Moreover, we carried out one training repetition of PreCNet-small, PreCNet-residual-error and PreCNet-single-LSTMs.

\begin{table}[!ht]
\caption{\textbf{Performance summary of 10 training repetitions on KITTI dataset.} Caltech Pedestrian Dataset was used for calculation of the values. See Section \ref{sec:quant_analysis} for details.}
\label{tab:kitti_repet}
\centering
\begin{tabular}{c||c|c|c}
\hline
 & MSE & PSNR & SSIM \\
\hline
\hline
best value  & 0.00205 & 28.4 & 0.929 \\
\hline
worst value  & 0.00220 & 28.1 &  0.928 \\
\hline
median & 0.00208 & 28.4 & 0.928 \\
\hline
\end{tabular}
\end{table}

We took the best model of 10 repetitions (according to SSIM) and compared it with state-of-the-art methods (see Table~\ref{tab:kitti_performance}). In the Table, the methods are sorted according to their SSIM values. If not stated otherwise, a network received ten input images and predicted the next one which was used during performance evaluation. Unless otherwise stated, the values were taken from the original articles. Values for BeyondMSE were taken from~\cite{liang2017dual}; values for DVF and CtrlGen were taken from~\cite{gao2019disentangling}. Values for PredNet were taken from~\cite{byeon2018contextvp}, because in~\cite{lotter2017deep} the values were averaged over nine (2-10) time steps. Values for ``PreCNet 7 input frames'' (see Table \ref{tab:input_length}), RC-GAN, DPRAConvLSTM and Lu et al. model were calculated after only seven, four, three and two input images (not ten), respectively. However, ``PreCNet 7 input frames'' and RC-GAN had better performance in this case than for input sequences of length ten. If this is true also for DPRAConvLSTM and Lu et al. model is not known. The number of parameters for DM-GAN and PredNet were taken from~\cite{byeon2018contextvp}.

\begin{table}[!ht]
\caption{\textbf{Next frame video prediction performance on Caltech Pedestrian Dataset after training on KITTI dataset.} See text for details.
}
\label{tab:kitti_performance}
\centering
\begin{tabular}{c|c|c|c|c}
\hline
 & \multicolumn{4}{c}{Caltech Pedestrian Dataset} \\
\hline
Method & MSE & PSNR & SSIM & \#param\\
\hline
\hline
Copy last frame & 0.00795 &  23.2 & 0.779 & -\\
\hline
BeyondMSE~\cite{mathieu2016deep} & 0.00326 & - & 0.881 & - \\
\hline
DVF~\cite{liu2017video} & - & 26.2 & 0.897 & - \\
\hline
DM-GAN~\cite{liang2017dual} & 0.00241 & - & 0.899 & 113M \\
\hline
CtrlGen~\cite{hao2018controllable} & - & 26.5 & 0.900 & - \\
\hline
PredNet~\cite{lotter2017deep} & 0.00242 & 27.6 & 0.905 & 6.9M\\
\hline
Lu et al.~\cite{lu2021video} & 0.00188 & 28.7 & 0.913 & 3.9M \\
\hline
RC-GAN~\cite{kwon2019predicting} & 0.00161 & 29.2 & 0.919 & - \\
\hline
ContextVP~\cite{byeon2018contextvp} & 0.00194 & 28.7 & 0.921 & 8.6M\\
\hline
DPG~\cite{gao2019disentangling} & - & 28.2 & 0.923 & - \\
\hline
CrevNet~\cite{Yu2020Efficient} & - & 29.3 & 0.925 & - \\
\hline
Jin et al.~\cite{jin2020exploring} & - & 29.1 & 0.927 & 7.6M \\
\hline
\bf{PreCNet (ours)} & 0.00205 & 28.4 & 0.929 & 7.6M\\
\hline
 \bf{PreCNet 7 input frames (ours)} & 0.00202 & 28.5 & 0.930 & 7.6M\\
\hline
 DPRAConvLSTM~\cite{yuan2021novel} & - & 30.2 & 0.930 & -\\
\hline
 IPRNN~\cite{chang2021iprnn} & \bf{0.00097} & \bf{31.0} & \bf{0.955} & -\\
\hline
\hline
PreCNet-small (ours)& 0.00220 & 28.0 & 0.919 & 0.8M\\
\hline
PreCNet-single-LSTMs (ours) & 0.00209 & 28.3 & 0.926 & 7.0M\\
\hline
 PreCNet-residual-error (ours) & 0.00212 & 28.2 & 0.927 & 5.6M\\
\end{tabular}
\end{table}

PreCNet achieved 2nd-3rd position in SSIM.
In MSE and PSNR, it was outperformed by four and seven other methods, respectively. The number of trainable parameters---for the models where it is available---is similar for all except  DM-GAN (113M) and PreCNet-small (0.8M). PreCNet with a small number of parameters still had comparable performance to other models and outperformed PredNet.
Replacement of the pairs of $\mathrm{convLSTM}^{up/down}_i$ by a single $\mathrm{convLSTM}_i$ and simplification of error blocks degraded the performance only slightly.

Moreover, we took the best trained model and investigated its performance for shorter input testing sequences. The results are shown in Table \ref{tab:input_length}. The network received input sequences with the given input length and predicted the next frame. Except the input sequence length, the experiential setting and the PreCNet network are the same as in Tab. \ref{tab:kitti_performance} (the values for input length 10 are the same as in Table \ref{tab:kitti_performance}). Copy last MSE is not the same for all input lengths because the test set was split into non-overlapping sequences with different lengths. 
\begin{table}[!ht]
\caption{\textbf{Next frame prediction performance for different length of input sequence.} See text for a description.}
\label{tab:input_length}
\centering
\begin{tabular}{c||c|c|c|c}
\hline
 & \multicolumn{4}{c}{Caltech Pedestrian Dataset} \\
\hline
Input length & MSE & PSNR & SSIM & Copy last MSE \\
\hline
\hline
3 & 0.00216 & 28.1 & 0.924 & 0.00796 \\
\hline
4 & 0.00208 & 28.4 & 0.928 & 0.00794 \\
\hline
5 & 0.00203 & 28.5 & 0.929 & 0.00795 \\
\hline
6 & 0.00204  & 28.5 & 0.930 & 0.00799 \\
\hline
7 & 0.00202  & 28.5 & 0.930 & 0.00798 \\
\hline
8 & 0.00203   & 28.5 & 0.930 & 0.00794 \\
\hline
9 & 0.00203  & 28.5 & 0.930 & 0.00794 \\
\hline
10 & 0.00205  & 28.4 & 0.929 & 0.00795 \\
\end{tabular}
\end{table}
The performance was significantly worse for input length 3 and slightly worse also for length 4. For longer sequences, it was stable. For input length 10, the performance even slightly decreased. A possible reason could be that the network was trained on sequences with length 10 and, therefore, it was not directly trained to predict the 11th frame after inputting 10 frames. To investigate the influence of sequence length during training, we trained a network with $l_s=5$---half of the basis sequence length---and evaluated it on the 6th frame after inputting 5 frames. We performed two repetitions with 1000 epochs and two repetitions with 2000 epochs. The results in all four cases were similar; SSIM was $0.924$ in all cases, PSNR varied from $28.2$ to $28.4$ and MSE was between $0.00209$ and $0.00215$. Therefore, the shorter sequence length during training degraded the performance.

As the training on \textbf{BDD100K dataset} required long training (large dataset), we performed only two training repetitions.
The performance is evaluated in Table \ref{tab:diff_datasets_perf}\footnote{Performance of the network from the other training repetition is: MSE 0.00169, PSNR 29.3, SSIM 0.938.
}.   
\begin{table}[!ht]
\caption{\textbf{Comparison of PreCNet performance on Caltech Pedestrian Dataset after training on KITTI (same as in Table \ref{tab:kitti_performance}) and BDD100K dataset} (see Section \ref{sec:exp_variants} for details).}
\label{tab:diff_datasets_perf}
\centering
\begin{tabular}{c|c|c||c|c|c}
\hline
 & \multicolumn{3}{c}{Caltech Pedestrian Dataset} \\
\hline
Training Set & \#frames & \#epochs & MSE & PSNR & SSIM \\
\hline
\hline
BDD100K  & 2M & 10000 & \bf{0.00167} & \bf{29.4} & \bf{0.938}\\
\hline
BDD100K  & 41K & 1000 & 0.00201 & 28.6 & 0.926\\
\hline
KITTI & 41K & 1000 & 0.00205 & 28.4 & 0.929\\
\hline
\end{tabular}
\end{table}
Usage of larger dataset led to significant performance improvement in all three measures. Comparing PreCNet trained on large BDD100K subset (2M) with the models trained on KITTI dataset (see Table~\ref{tab:kitti_performance}), our model was second in SSIM and third in PSNR and MSE.

In order to evaluate effect of different properties of BDD100K and KITTI datasets on performance, we created a small version of the BDD100K dataset with only approx. 41K frames (similar size as the size of KITTI) and used the same training parameters which were used for training on KITTI. The performance on this dataset was similar to performance on KITTI\footnote{We performed 3 training repetitions on BDD100K with 41K frames. In Table~\ref{tab:diff_datasets_perf}, there is performance of the best one (according to SSIM). Performance of the other two is MSE \{0.00199; 0.00202\}, SSIM \{0.925; 0.926\}, PSNR \{28.6; 28.6\}.}. This suggests that the ``quality'' of the training set (BDD100K vs. KITTI) is not the key factor for obtaining better performance in this case.
We studied the effect of the number of training epochs as well. Validation loss on the small subset of BDD100K (41K frames) started to increase during training (1K epochs), indicating overfitting. Thus, we can exclude the possibility that training for 10K epochs would further improve performance. Hence, we claim that it is really the dataset size that is the enabling factor for performance and that permitted the results obtained for BDD100K (2M frames, 10K epochs). 

\subsubsection{Qualitative analysis}
\begin{figure*}[!ht]
\centering
\includegraphics[width=7in]{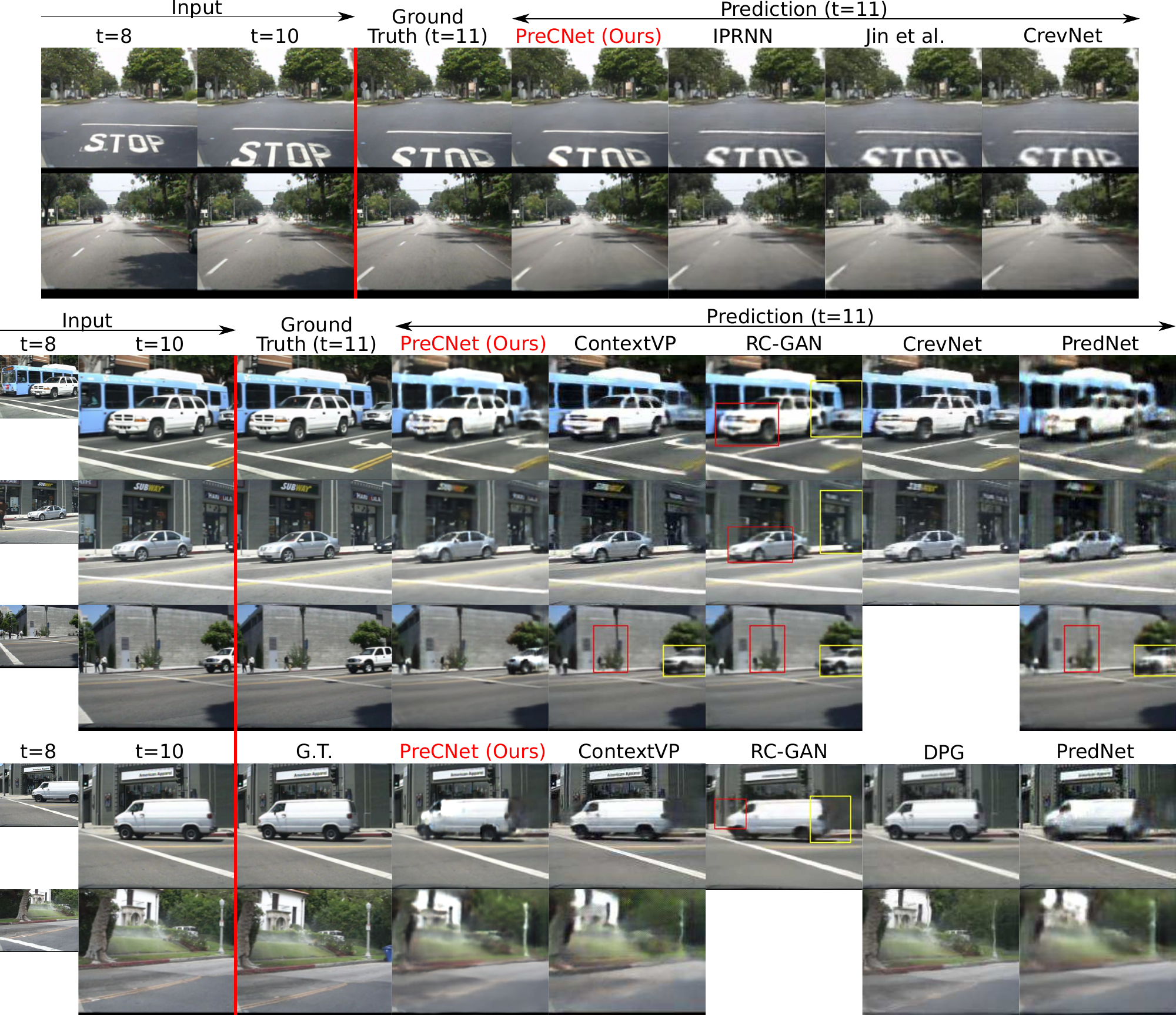}
\caption{\textbf{Qualitative comparison of PreCNet with others state-of-the-art methods on Caltech Pedestrian Dataset.} All models were trained on KITTI dataset. Ten input frames were given (see frames for $t=8, t=10$), the next one ($t=11$) was predicted (RC-GAN used only four input frames -- see Section \ref{sec:quant_analysis} for explanation) by the models (for references see Table \ref{tab:kitti_performance}). The images of predictions of other models are copied from original or other cited papers
(see references in Table \ref{tab:kitti_performance}). Position of the sequences in Caltech Pedestrian Dataset by rows; set06-v013, set06-v000, set07-v011, set10-v010, set10-v009, set10-v010, set06-v009.}  
\label{fig:kitti_comparison}
\end{figure*}

In Fig.~\ref{fig:kitti_comparison}, there is a qualitative comparison of PreCNet with other state-of-the-art methods trained on KITTI dataset (see Table \ref{tab:kitti_performance}). The way of obtaining the predicted frames used for the analysis is the same as for Quantitative analysis (see the predicted frame at $t=11$ in Fig.~\ref{fig:next_frame_pred}). For a qualitative comparison of PreCNet with the model by Jin et al.~\cite{jin2020exploring} see Fig.~\ref{fig:jin_multiple} (the predicted frame at $t=11$). 

To assess which of the methods is best through visual inspection is not straightforward; none of the models is better than the others in all aspects and shown frames (excluding PredNet which produced significantly worse predictions). For example, in the last row of Fig.~\ref{fig:kitti_comparison}, DPG has generally the sharpest prediction but PreCNet predicted the street lamp significantly better.
To compare our model with the IPRNN, which significantly outperformed all models in all metrics used, we used the sequences from the IPRNN article \cite{chang2021iprnn}. On these sequences (see Fig. \ref{fig:kitti_comparison}, first two rows), there is no apparent qualitative difference between the predictions by IPRNN and PreCNet. For example, ``STOP'' sign in the sequence from the first row is predicted sharper by PreCNet than by IPRNN and the other models.

\begin{figure}[!ht]
\centering
\includegraphics[width=3.5in]{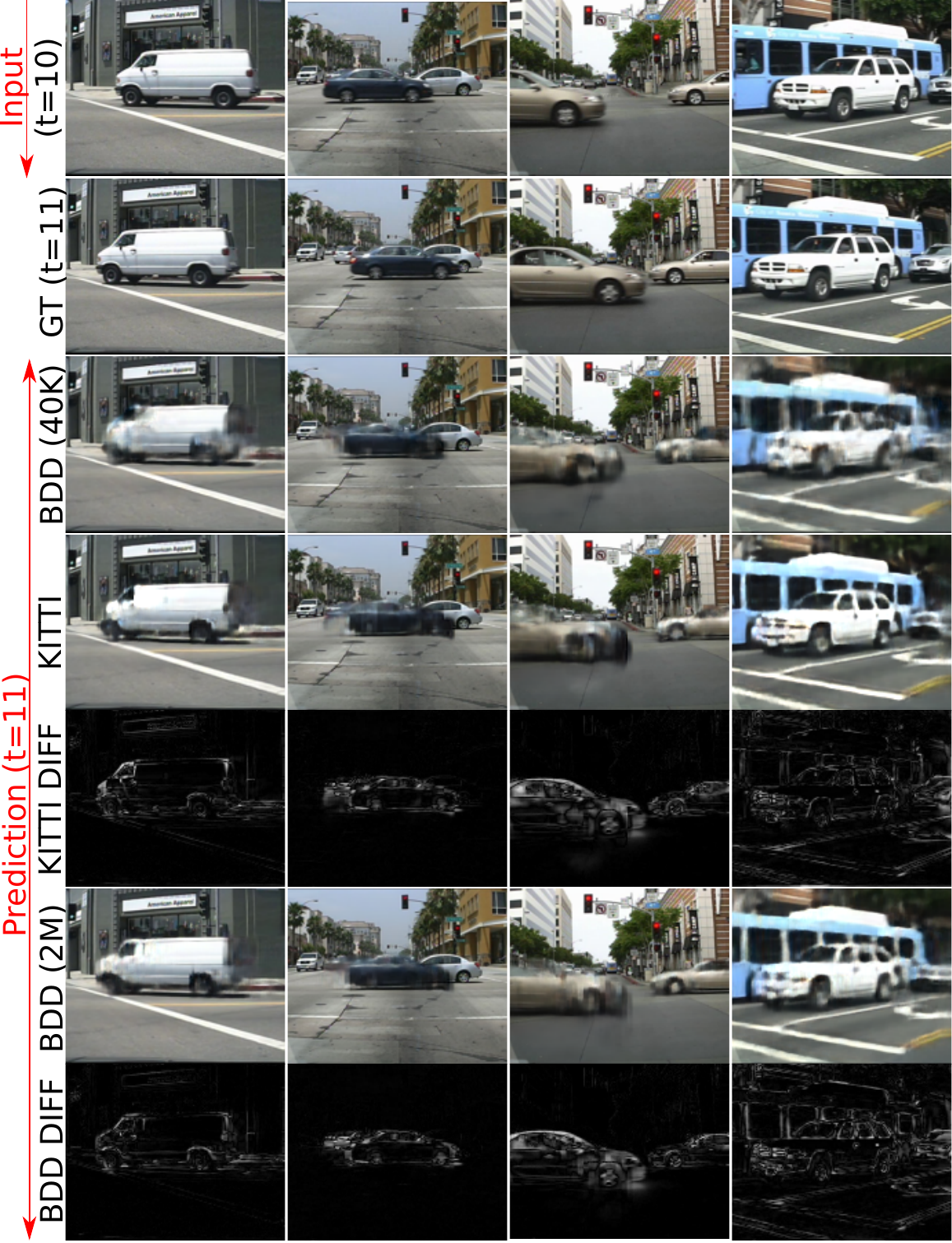}
\caption{\textbf{Qualitative comparison of PreCNet performance on Caltech Pedestrian Dataset after different training variants.} First row corresponds to the last frame of the input sequence with length 10. Second row corresponds to the ground truth frame. Next rows correspond to the predicted frames of different models which correspond to the models from quantitative evaluation in Table \ref{tab:diff_datasets_perf} and related residual images -- difference between the predicted and the actual frame.  Position of the sequences in Caltech Pedestrian Dataset by columns; set10-v010, set06-v001, set07-v011, set07-v011. In contrast with Fig.~\ref{fig:kitti_comparison}, the meaning of horizontal and vertical arrangement is inverted. To see whole input sequences and related predictions check \textit{Supplementary materials -- Examples of next frame video prediction sequences}.}   
\label{fig:diff_datasets_perf}
\end{figure}

In Fig.~\ref{fig:diff_datasets_perf}, KITTI and BDD100K (both 2M and 41K) training variants (see Table \ref{tab:diff_datasets_perf}) are compared. Usage of large BDD100K dataset (with approx. 2M frames) for training led to significant improvement of all the measures (see Table \ref{tab:diff_datasets_perf}) in comparison with training on KITTI dataset.    
It manifested also in the visual quality of prediction of fast moving cars as you can see in the second and third columns of the figure. The phantom parts of the predicted cars were reduced. It also led to better shapes of the predicted cars as you can see in the prediction in the first column (focus on the front part of the van). On the other hand, in some cases training on BDD100K dataset led to blurrier predictions than training on KITTI (see the last column).

\subsection{Multiple frame prediction}
\label{sec:multiple_frame_pred}
\begin{figure*}[!ht]
\centering
\includegraphics[width=7in]{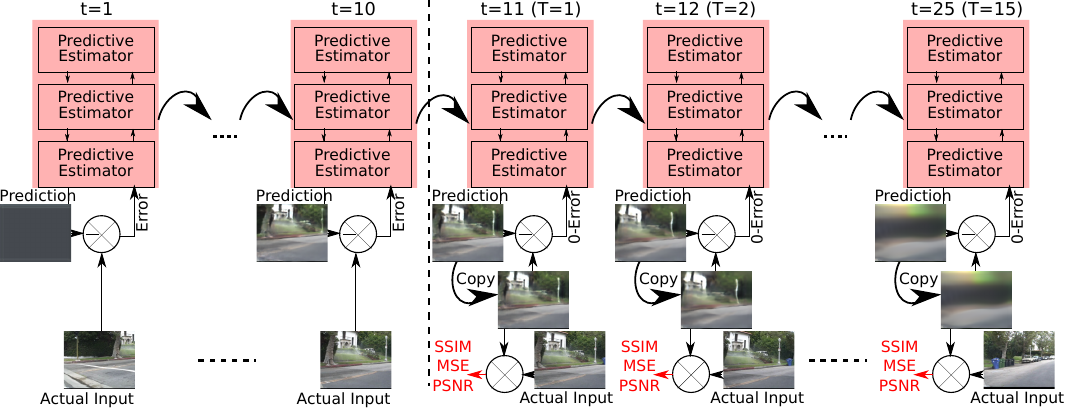}
\caption{\textbf{Multiple frame video prediction evaluation schema.} After inputting 10 frames, the predicted frames are inputted instead of the actual frames. The prediction errors are therefore zeros.
The predicted frames are compared---using MSE, PSNR, SSIM---with the actual inputs. We used this schema for both quantitative and qualitative analysis of Multiple frame prediction (see Section \ref{sec:multiple_frame_pred}).
}  
\label{fig:multiple_frame_pred}
\end{figure*}

For multiple frame prediction, we used the same trained models which we used for next frame video prediction (see Section \ref{sec:next_frame_pred}). The network had access to the first 10 frames---same as in next frame video prediction. Then, in each timestep, the network produced next frame and this next frame was used as the actual input (as illustrated in Fig.~\ref{fig:multiple_frame_pred}). 
Therefore, the prediction error between the prediction and input frame was zero.

We briefly explored fine-tuning of the network for multiple frame prediction \cite{lotter2017deep}. To generate multiple future frames, the predicted frames---produced by the pre-trained network for next frame prediction---were used as inputs after inputting 10 frames. The network was trained to minimize the mean absolute error between the predicted and ground-truth frames. According to our preliminary results, this did not significantly improve multiple frame prediction performance.

Please note the different meaning of timestep labels $t$ and $T$: small $t$ starts at the beginning of a sequence, in contrast with capital $T$, which starts at the beginning of a predicted sequence (see the timestep labels in Fig.~\ref{fig:multiple_frame_pred}).
Code needed for generation of the results presented is publicly available~\cite{github-precnet}.  

\subsubsection{Quantitative results}
In Table \ref{tab:multiple}, there is a quantitative comparison of PreCNet, PredNet, CrevNet and RC-GAN for multiple frame prediction. The methods obtained sequences with a fixed length (10 for PredNet, CrevNet and PreCNet; 4 for RC-GAN; see Section \ref{sec:quant_analysis} for explanation) of Caltech Pedestrian Dataset and outputted predictions 15 steps ahead (CrevNet only 12). CrevNet, RC-GAN, PredNet and PreCNet (KITTI) were trained on KITTI. PreCNet was also trained on a subset of BDD100K with size 2M. Values for PredNet and RC-GAN were copied from~\cite{kwon2019predicting}. Values for CrevNet were taken from~\cite{Yu2020Efficient}. Some values for $T=1$ from Tables~\ref{tab:kitti_performance} and~\ref{tab:diff_datasets_perf} are slightly different because the test set used there was split into non-overlapping sequences with different length (11 vs. 25).

For SSIM, PreCNet trained on KITTI outperformed PredNet until timestep $T=9$ ($t=19$)
when the values became equal and then PreCNet started to lose. For PSNR, PreCNet started to lose earlier ($T=6$).
RC-GAN and CrevNet outperformed PreCNet in nearly all timesteps for SSIM\footnote{In $T=1$, SSIM for PreCNet was $0.930$ and for CrevNet $0.925$.} and RC-GAN also in all timesteps for PSNR.

\begin{table}[!ht]
\caption{\textbf{A quantitative comparison of selected methods for multiple frame prediction.} See text for a description.}
\label{tab:multiple}
\centering
\begin{tabular}{c|c|c c c c c c}
\hline
Method & & T=1 & 3 & 6 & 9 & 12 & 15 \\
\hline
\hline
\multirow{2}{*}{PredNet~\cite{lotter2017deep}} & PSNR & 27.6 & 21.7 & 20.3 & 19.1 & 18.3 & 17.5\\
& SSIM & 0.90 & 0.72 & 0.66 & 0.61 & 0.58 & 0.54\\
\hline
\multirow{2}{*}{RC-GAN~\cite{kwon2019predicting}} & PSNR & 29.2 & 25.9 & 22.3 & 20.5 & 19.3 & 18.4\\
& SSIM & 0.91 & 0.83 & 0.73 & 0.67 & 0.63 & 0.60\\
\hline
\multirow{1}{*}{CrevNet~\cite{Yu2020Efficient}} 
& SSIM & 0.93 & 0.84 & 0.76 & 0.70 & 0.65 & - \\
\hline
PreCNet & PSNR & 28.5 & 23.4 & 20.2 & 18.4 & 17.2 & 16.3\\
(KITTI) & SSIM & 0.93 & 0.82 & 0.69 & 0.61 & 0.56 & 0.53\\
\hline
\hline
PreCNet & PSNR & 29.5 & 24.6 & 21.4 & 19.4 & 18.3 & 17.4\\
(BDD100K 2M)& SSIM & 0.94 & 0.85 & 0.73 & 0.65 & 0.59 & 0.56\\
\end{tabular}
\end{table}

We also added PreCNet trained on the large subset of BDD100K to the comparison. Then PreCNet outperformed PredNet in all timesteps for SSIM and most timesteps for PSNR; in timestep $T=15$ it reversed. However, CrevNet and RC-GAN still outperformed PreCNet in most timesteps. For SSIM, PreCNet had better results than CrevNet and RC-GAN only for predicted frames in $T\in\{1, 3\}$. For PSNR, RC-GAN was outperformed by PreCNet only for $T=1$. 

In summary, PreCNet started with mostly better predictions than its competitors, however, its performance tended to degrade faster for prediction further ahead. 

\subsubsection{Qualitative analysis}
\label{sec:multiple_frame_qualit_an}
The methods were compared using the sequences used in \cite{kwon2019predicting}. Moreover, PreCNet was separately compared to the model by Jin et al.  (Fig.~\ref{fig:jin_multiple}).
Fig.~\ref{fig:multiple} provides one example (for another illustration, see \textit{Supplementary Materials -- Multiple frame video prediction sequence}).
\begin{figure*}[!ht]
\centering
\includegraphics[width=7in]{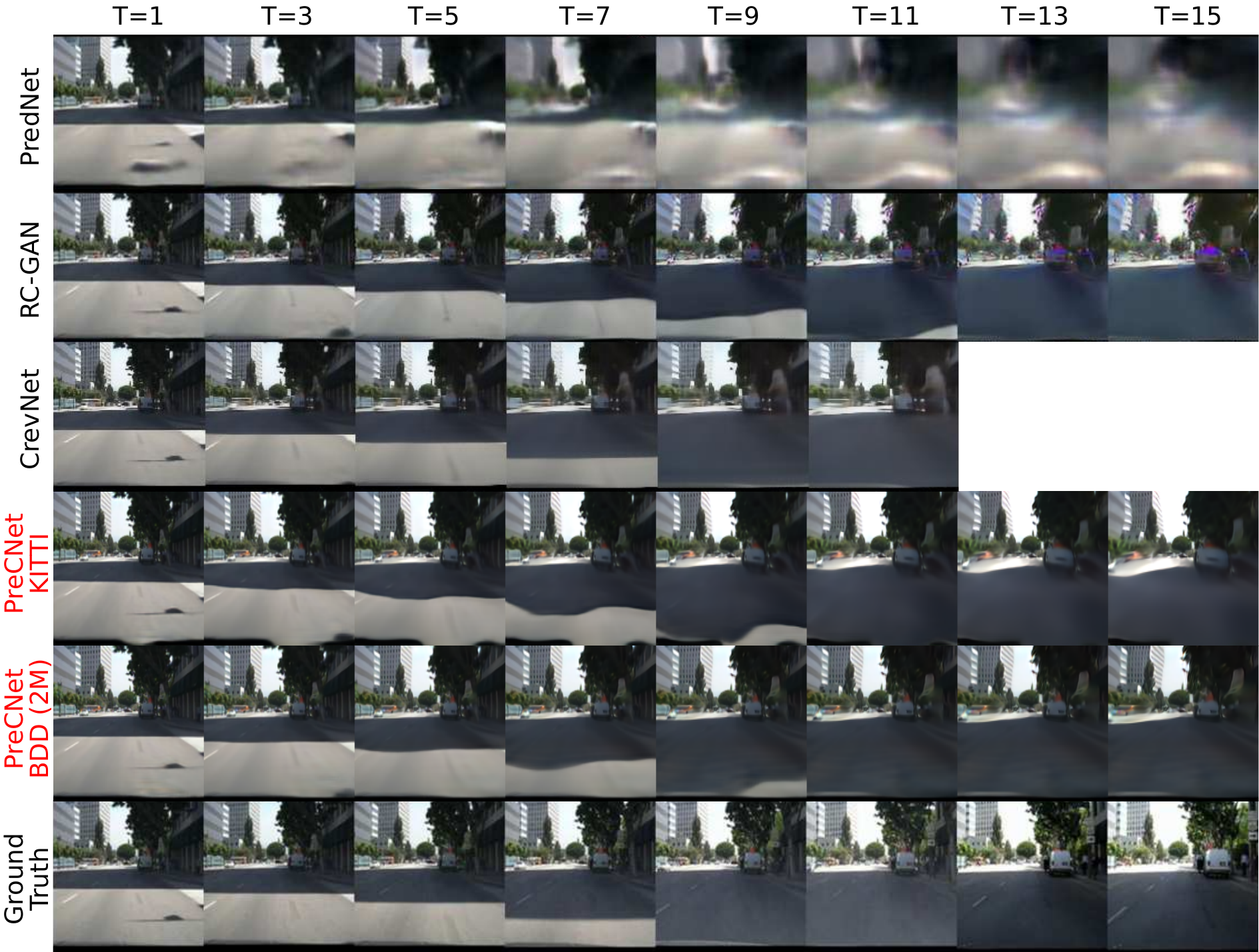}
\caption{\textbf{A qualitative comparison of selected methods for multiple frame prediction.} The methods obtained sequence with fixed length (10 for PredNet, CrevNet and PreCNet, 4 for RC-GAN; see Section \ref{sec:quant_analysis} for explanation) of Caltech Pedestrian Dataset and outputted predictions 15 steps ahead. RC-GAN, PredNet, CrevNet and PreCNet (KITTI) were trained on KITTI. PreCNet was also trained on subset of BDD100K with size 2M. This should be noticed during comparison with the other four trained models. This figure was obtained from the figure from~\cite{kwon2019predicting} by adding sequences for PreCNet and CrevNet (taken from~\cite{Yu2020Efficient}). Location of the sequence in Caltech Pedestrian Dataset is set10-v009. 
Another qualitative comparison (without CrevNet), with different sequence, is in \textit{Supplementary materials -- Multiple frame video prediction sequence}.
}
\label{fig:multiple}
\end{figure*}
Predictions by PreCNet appear less blurred than those by PredNet and by the model of Jin et al. (see Fig.~\ref{fig:jin_multiple}). This is especially apparent for the later predicted frames.
Compared to RC-GAN, predicted frames by PreCNet trained on KITTI seem to have more natural colors and background is mostly less blurred (focus on the buildings in the background). PreCNet trained on large subset of BDD100K (2M of frames) produced even less blurred frames.
Comparison with CrevNet is not straightforward. For example, CrevNet captured the geometry of the shadow of the building on the road better than PreCNet. On the other hand, it produced a phantom object (see right side of the road in timesteps $9, 11$) which is not present (or negligible) in the corresponding frames by PreCNet.

\begin{figure}[!ht]
\centering
\includegraphics[width=3.49in]{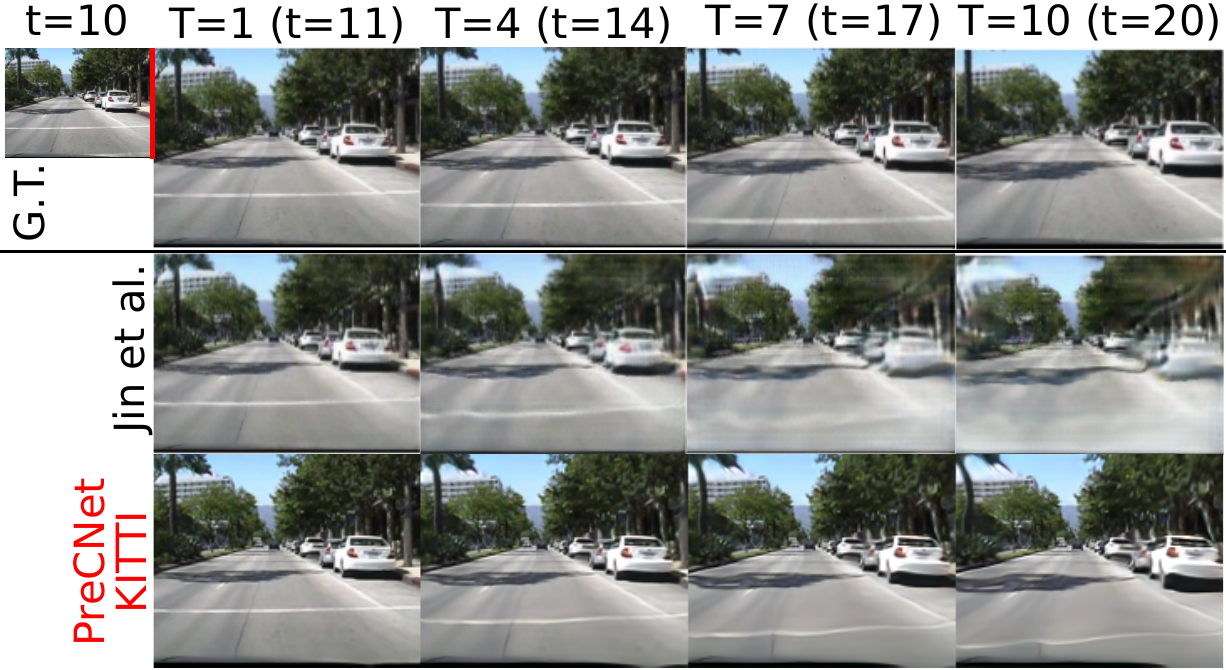}
\caption{\textbf{A qualitative comparison of PreCNet and model by Jin et al.~\cite{jin2020exploring} for multiple frame prediction.} The sequence is from Caltech Pedestrian Dataset and was used in~\cite{jin2020exploring}. Both models were trained on KITTI dataset. Input and output sequence lengths were 10. Location of the sequence in Caltech Pedestrian Dataset is set05-v012. As this sequence was used in~\cite{jin2020exploring} without downsampling to 10 fps, which is in contrast to other usage of Caltech Pedestrian Dataset in this paper, we left it without downsampling as well.
}
\label{fig:jin_multiple}
\end{figure}

\section{Conclusion, Discussion, Future Work}

In this work, the seminal predictive coding model of Rao and Ballard~\cite{rao1999predictive}---here referred to as \textit{predictive coding schema}---has been cast into a modern deep learning framework, while remaining as faithful as possible to the original schema. The similarities and differences are elaborated in detail. We also claim and explain that the network we propose (PreCNet) is more congruent with \cite{rao1999predictive} than others based on the deep learning framework that take inspiration from predictive coding; the case of PredNet~\cite{lotter2017deep} is studied explicitly. 
PreCNet was tested on a widely used next frame video prediction benchmark---KITTI for training (41k images), Caltech Pedestrian Dataset for testing---, which consists of images from an urban environment recorded from a car-mounted camera. On this benchmark, we outperformed most of the state-of-the-art methods and achieved 2nd-3rd rank when measured with the Structural Similarity Index (SSIM)---a performance measure that should best correlate with human perception.
Performance on all three measures was further improved when a larger training set (2M images from BDD100k; to our knowledge, biggest dataset ever used in this context) was employed. This may suggest that the current practice based on the rather small KITTI dataset used for training may be limiting in the long run. At the same time, the task itself seems highly relevant, as virtually unlimited amount of data and without any need for labeling is readily available.

Below, we discuss some limitations of this work. For some fast moving objects in the scene, PreCNet could not restore their structure precisely (see e.g., the third column of Fig.~\ref{fig:diff_datasets_perf} where the car contours are not preserved). This may be a drawback of the cost function that minimizes the per-pixel loss. Perceptual loss (e.g., \cite{johnson2016perceptual}) based on high-level feature differences between frames might alleviate this problem.

In multiple frame video prediction, qualitatively, the frames predicted by PreCNet look reasonable and in some aspects better than some of the competitors. However, a quantitative comparison reveals that PreCNet performance degrades slightly faster than that of its competitors when predicting up to 15 frames ahead. 
We speculate that architectures which achieve multiple frame prediction by recurrent feeding of previous predictions may not achieve their best performance for next and multiple frame predictions at the same time. Increasing performance for multiple frame prediction may decrease performance for next frame prediction and vice versa. We performed fine-tuning of our network for multiple frame prediction, but preliminary results did not show any significant improvement. PreCNet and predictive coding in general is perhaps intrinsically more suited for next frame prediction. This remains to be further analyzed.

In the future, we plan to analyze the representations formed by the proposed network. It would be interesting to study how much of the semantics of the urban scene has the network ``understood'' and how that is encoded. For example, our network has not quite figured out that every car has a finite length and its end should be predicted at some point when it is not occluded anymore. In our model, best results on the task were achieved when only prediction error on the bottom level---difference between the actual frame and the predicted one---was minimized during learning. Rao and Ballard~\cite{rao1999predictive}, on the other hand, minimized this error on every level of the network hierarchy, which may have an impact on the representations formed.
Testing on a different task, like human action recognition (e.g., \cite{liang2017dual,vondrick2016generating,lotter2017deep}) is also a possibility. Finally, some datasets feature also other signals apart from the video stream. Adding inertial sensor signals or the car's steering wheel angle or throttle level is another avenue for future research.

We want to close with a discussion of the implications of our model for neuroscience. Casting the \textit{predictive coding schema} into a deep learning framework has led to exceptional performance on a contemporary task, without being explicitly designed for it. In the future, we plan to analyze the consequences for computational neuroscience. While receptive field properties in sensory cortices remain an active research area (e.g., \cite{singer2018sensory}), a question remains whether the deep learning approach can lead to a better model than, for example, that of Rao and Ballard~\cite{rao1999predictive}. Richards et al.~\cite{richards2019deep} and Lindsay~\cite{lindsay2020convolutional} provide recent surveys of this perspective. An investigation of this kind has recently been performed for PredNet~\cite{lotter2020neural}.

\appendices

\ifCLASSOPTIONcompsoc
  \section*{Acknowledgments}
\else
  \section*{Acknowledgment}
\fi
Z.S. and M.H. were supported by the Czech Science Foundation (GA \v{C}R), project no. 20-24186X.
T.S. acknowledges the support of the OP VVV MEYS funded project
CZ.02.1.01/0.0/0.0/16\_019/0000765 ``Research Center for Informatics''. The access to the computational infrastructure available through this project is also gratefully acknowledged. We would like to thank to the authors of PredNet~\cite{lotter2017deep} for making their source code public which significantly accelerated the development of PreCNet. 

\ifCLASSOPTIONcaptionsoff
  \newpage
\fi

\bibliographystyle{IEEEtran}
\bibliography{IEEEabrv,ref}

\vskip -2\baselineskip plus -1fil

\begin{IEEEbiography}[{\includegraphics[width=1in,height=1.25in,clip,keepaspectratio]{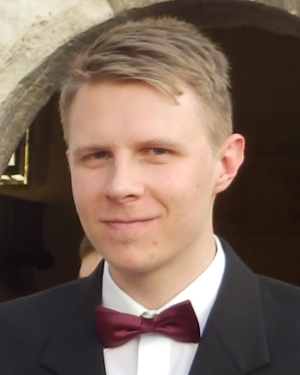}}]{Zdenek Straka}
received the BS degree (summa cum laude) in Cybernetics \& Robotics and the MS degree (summa cum laude) in Artificial Intelligence from the Faculty of Electrical Engineering, Czech Technical University in Prague, Czech Republic, in 2014 and 2016, respectively, where he is currently pursuing the Ph.D. degree with the Humanoid and Cognitive Robotics group. His current research interests include neural networks, neurorobotics, computational neuroscience and peripersonal space representations. He received the ENNS Best Paper Award at ICANN 2017.
\end{IEEEbiography}
\vskip -2\baselineskip plus -1fil
\begin{IEEEbiography}[{\includegraphics[width=1in,height=1.25in,clip,keepaspectratio]{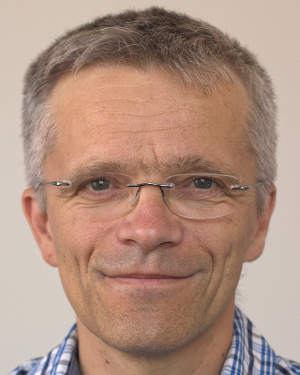}}]{Tom{\'a}{\v s} Svoboda}
received the Ph.D. degree in Artificial Intelligence and Biocybernetics from the Czech Technical University in Prague, Czech Republic, in 2000. He spent three postdoctoral years with the Computer Vision Group, ETH Zurich, Switzerland. He is a Full Professor and Chair of the Department of Cybernetics, FEE, CTU, the Director of Cybernetics and Robotics PhD study program, and he is also on the Board of the Open Informatics programme. He has published articles on multi-camera systems, omnidirectional cameras, image-based retrieval, learnable detection methods, and USAR robotics, he led CTU-CRAS-NORLAB team within DARPA SubT Challenge.  His research interests include multimodal perception for autonomous systems, machine learning for better simulation and robot control, and related applications in the automotive industry and robotics.
\end{IEEEbiography}
\vskip -2\baselineskip plus -1fil

\begin{IEEEbiography}[{\includegraphics[width=1in,height=1.25in,clip,keepaspectratio]{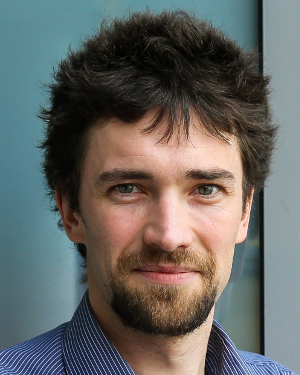}}]{Matej Hoffmann} completed the PhD degree and then served as Senior Research Associate at the Artificial Intelligence Laboratory, University of Zurich, Switzerland (Prof. Rolf Pfeifer, 2006--2013). In 2013 he joined the iCub Facility of the Italian Institute of Technology (Prof. Giorgio Metta), supported by a Marie Curie Intra-European Fellowship. In 2017, he joined the Department of Cybernetics, FEE, CTU in Prague, where he is currently serving as Associate Professor and Coordinator of the Humanoid and Cognitive Robotics group. His research interests are in humanoid, cognitive developmental, and collaborative robotics.

\end{IEEEbiography}

\end{document}


\maketitle

These are Supplementary materials to the article \textit{PreCNet: Next Frame Video Prediction Based on Predictive Coding}. They are structured as follows: (i) Network and training parameters summary and details, (ii) PreCNet and PredNet schema transformation, (iii) examples of next frame video prediction sequences, (iv) a figure with qualitative comparison of multiple frame video prediction using different methods.

\section{Network and training parameters summary and details}
To train a neural network, it is necessary to define the structure and hyperparameters of the network, the loss function and parameters of the optimization method for gradient descent\footnote{Notice that this is sufficient for gradient descent calculation and modern deep learning frameworks can perform this calculation automatically.}. In this section, there is a summary and details of all parameters. 

\subsection{Loss function parameters}
The loss was defined in subsection ``Training of the model'' as
\begin{equation}
\label{eq:Ltrain_supp}
L_{\mathit{train}}=\sum_{m=1}^{M}L_{\mathit{seq}}(m),    
\end{equation}
where $L_{\mathit{seq}}(j)$ is loss related to $j^{th}$ sequence, $M$ is the number of image sequences. The loss of sequence is defined as 
\begin{equation}
\label{eq:Lseq_supp}
L_{\mathit{seq}} (j)=\sum^{l_s}_{t=1} \mu_t \sum^N_{l=0} \frac{\lambda_l}{n_l} \sum_{i=1}^{n_l} E^t_l(i).    
\end{equation}
$E_l^t(i)$ is the error of the $i^{th}$ unit in the module $l$ at time $t$ (see Algorithm 1 for calculation description).
For the main experiments, the parameters are summarized (see subsection ``Next frame video prediction'' for details) in the following table.
\begin{center}
\begin{tabular}{ |c|c| } 
 \hline
 parameter & description  \\ 
 \hline
 $l_s=10$ & length of the sequence  \\ 
 $N=2$ & number of the highest module \\
 $\mu_{t=1}=0$; $\mu_t=\frac{1}{l_s-1}=\frac{1}{9}$ for $t \in \{2,\dots, 10\}$ & time weighting factor \\
 $\lambda_0=1, \lambda_{1,2}=0$ & module weighting factors  \\ 
 $n_0= 128 \cdot 160 \cdot 6=122880, n_1=614400, n_2=307200$ & number of error units in the modules  \\
 \hline
\end{tabular}
\end{center}

\subsection{Network hyperparameters}
\begin{sidewaysfigure}
\centering
\includegraphics[width=8.5in]{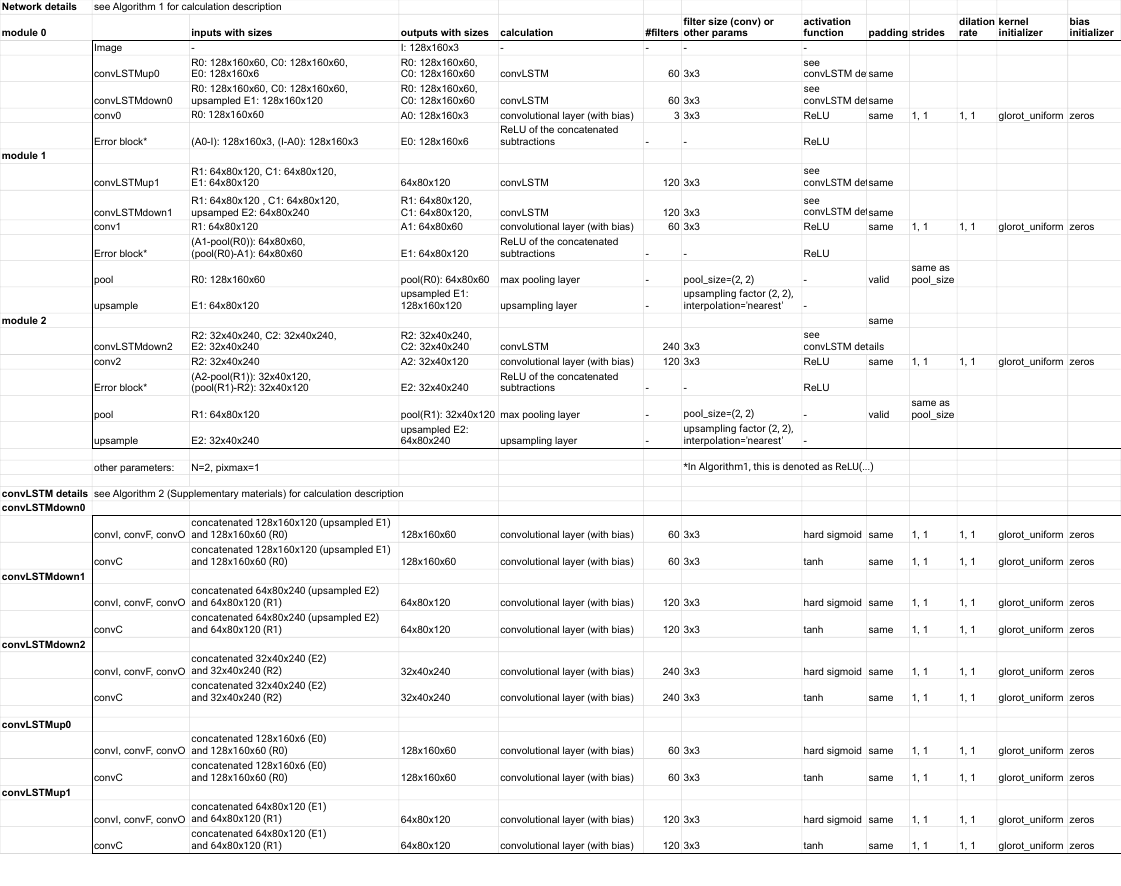}
\caption{\textbf{Network hyperparameters summary}}
\label{fig:netw_details_table}
\end{sidewaysfigure}

Network hyperparameters with details are in Fig. \ref{fig:netw_details_table} (this document). See Algorithm 1 and 2 for the corresponding calculation description.

\subsection{Optimization parameters}
To minimize the loss (\ref{eq:Ltrain_supp}), stochastic gradient descent optimization method Adam\cite{adam2015} with the following parameters was used (see subsection ``Training parameters'' for details).
\begin{center}
\begin{tabular}{ |c|p{60mm}| } 
 \hline
 parameter & description  \\ 
 \hline
 $\#epoch=1000/10000$ (\{KITTI, BDD100K 41K\}/\{BDD100K 2M\}) & number of training epochs  \\

 \hline
$bs=4$ & batch size \\
 \hline
   $lr=0.001/0.0005$ (\{KITTI, BDD100K 41K\}/\{BDD100K 2M\}) & learning rate for first 900/9900 epochs  \\
 \hline
   $lr=0.0001$ & learning rate for last 100 epochs  \\
 \hline
 training set: $500$, validation: $100$ sequences & number of randomly selected sequences in one epoch \\ 
  \hline
 $\beta_1=0.9, \beta_2=0.999$ & Exponential decay rates for the $1^{st}$ and $2^{nd}$ moment estimates. \\
  \hline
mean absolute error & optimizer loss used to minimize the distance between loss (\ref{eq:Lseq_supp}) and desired zero value (see the training script for details \cite{github-precnet})\\
 \hline
\end{tabular}
\end{center}

\subsection{convLSTM}
A convolutional LSTM (convLSTM) is a neural network architecture \cite{xingjian2015convolutional} which combines LSTM \cite{hochreiter1997long} and convolutional neural networks. The convLSTM, which we used, follows dynamics from the ``No peepholes'' variant of LSTM \cite{greff2016lstm}. This LSTM variant has lower computational cost without significant performance decrease \cite{greff2016lstm}. It enabled a better comparison with PredNet which used the same convLSTM variant.    

The calculation $\mathbf{convLSTM_l^{dir}}$, where $\mathbf{l} \in \{0, 1, \dots N\}$, $\mathbf{dir} \in \{down, up\}$ and $N$ is the number of the highest module, is shown below. Weights are denoted by the symbol $W$ and biases by $b$. Symbol ``$\ast$'' represents the convolutional operator and ``$\circ$'' denotes element-wise product. As the activation functions, hard sigmoid $\sigma$ and hyperbolic tangent $\tanh$ are used. Hidden, error and cell states $R_{input}, E_{input}, C_{input}$ are inputs of the calculation.
\begin{gather*}
\mathbf{convLSTM_{l}^{dir} (}R_{input}, C_{input}, E_{input} \mathbf{):} \\ 
I  \leftarrow  \sigma(W_{EI}^{dir,l} \ast E_{input} + W_{RI}^{dir,l} \ast R_{input} + b_I ) \\
F \leftarrow  \sigma(W_{EF}^{dir,l} \ast E_{input} + W_{RF}^{dir,l} \ast R_{input} + b_F ) \\
O \leftarrow \sigma(W_{EO}^{dir,l} \ast E_{input} + W_{RO}^{dir,l} \ast R_{input} + b_O ) \\
C \leftarrow F \circ C_{input} + I \circ \tanh(W_{EC}^{dir,l} \ast E_{input} + W_{RC}^{dir,l} \ast R_{input} + b_C) \\
R \leftarrow O \circ \tanh(C) \\  
\mathbf{return} \ R, C
\end{gather*}
In Algorithm \ref{alg:convLSTM}, it is rewritten to a more concise form, which is also very close to the actual implementation. For details see Fig. \ref{fig:netw_details_table} in this document.
\stepcounter{algorithm}
\begin{algorithm}
\caption{For given $l \in \{0, 1, \dots N\}$, $dir \in \{down, up\}$ and inputs $E_{input}, R_{input}, C_{input}$, $\textrm{convLSTM}_l^{dir}$ calculation is performed. Convolutional layers $\textrm{convI}_l^{dir}, \textrm{convF}_l^{dir}, \textrm{convO}_l^{dir}, \textrm{convC}_l^{dir}$ with biases are used during the calculation. Activation function for $\textrm{convI}_l^{dir}, \textrm{convF}_l^{dir}, \textrm{convO}_l^{dir}$ is hard sigmoid, for $\textrm{convC}_l^{dir}$ it is $\tanh$. Output states are $C$ and $R$. Concatenations of states $E_{input}$ and $R_{input}$ is denoted by putting them into curly brackets $\{E_{input}, R_{input}\}$.}

\label{alg:convLSTM}
\begin{algorithmic}
\STATE
$\mathbf{convLSTM_{l}^{dir} (}R_{input}, C_{input}, E_{input} \mathbf{):}$ \\
$I \leftarrow \textrm{convI}_l^{dir}(\{E_{input},R_{input}\})$ \\
$F \leftarrow \textrm{convF}_l^{dir}(\{E_{input},R_{input}\})$ \\
$O \leftarrow \textrm{convO}_l^{dir}(\{E_{input},R_{input}\})$ \\
$C \leftarrow F \circ C_{input} + I \circ \textrm{convC}_l^{dir}(\{E_{input},R_{input}\})$ \\
$R \leftarrow O \circ \tanh(C)$ \\
Return $R, C$
\end{algorithmic}
\end{algorithm}

\section{Schema transformation}

Transformation between two different ways of schema visualization. See Fig.~3 in the article for an explanation.
\begin{figure}[H]
\centering
\includegraphics[width=7in]{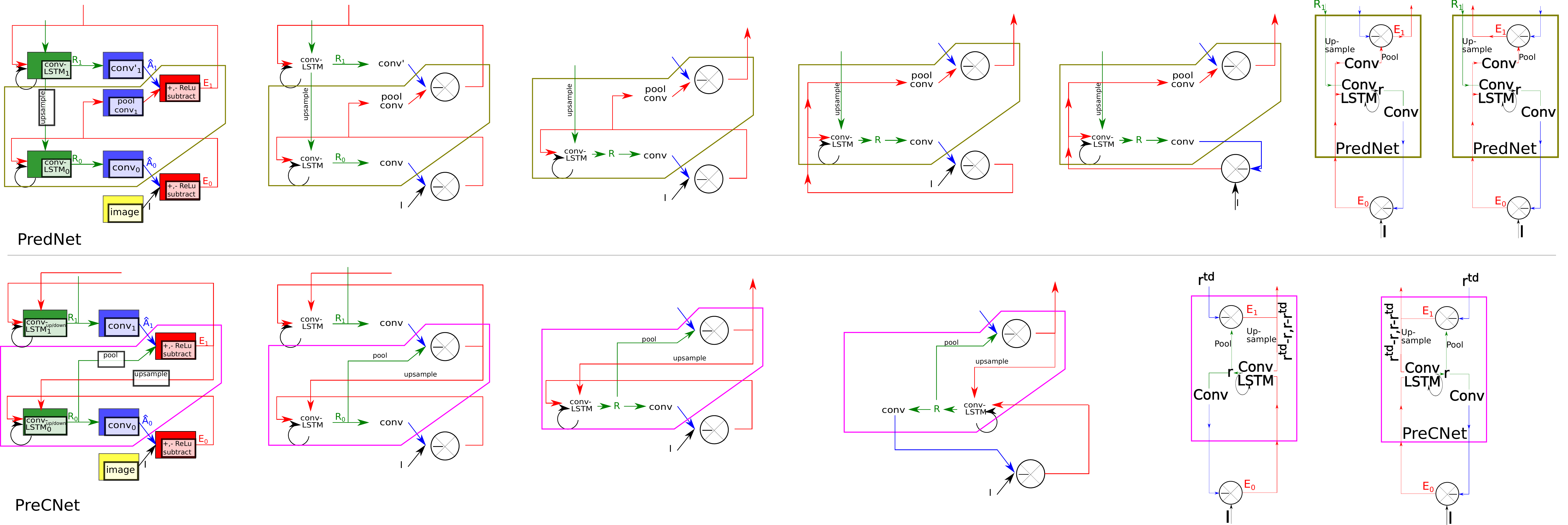}
\end{figure}

\section{Examples of next frame video prediction sequences}
Examples of sequences with next predicted frames are presented bellow. Last predicted frames (timestep 11) from the following figures were used for qualitative analysis (Fig.~6) in the article (see the figure caption for details). First row contains actual images, the second contains PreCNet's one step ahead predictions, for KITTI and BDD100K (2M), the third contains residual images (difference between the prediction and the actual image). For explanation of the generation process see Fig.~4.  
\subsection{PreCNet trained on KITTI}

\begin{figure}[H]
\centering
\includegraphics[width=7in]{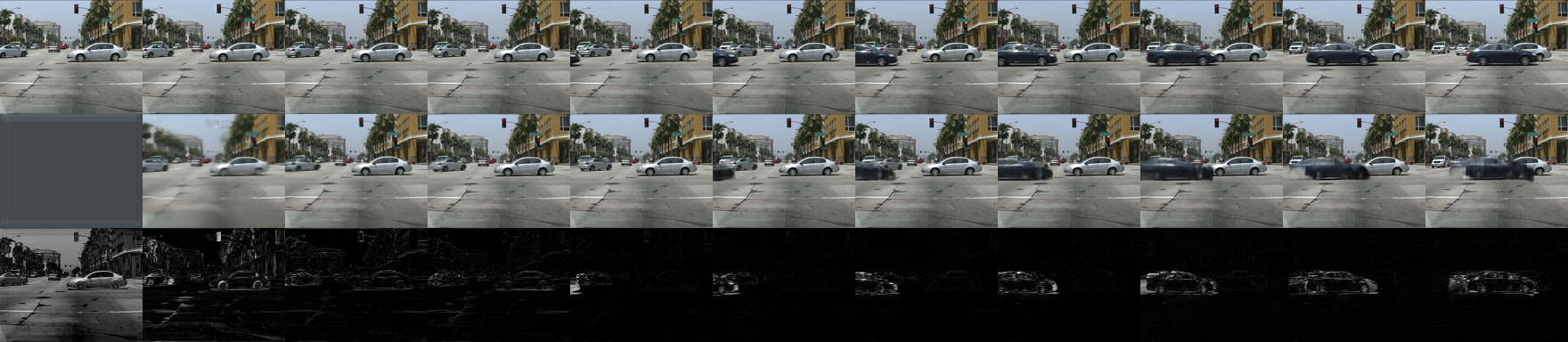}
\end{figure}

\begin{figure}[H]
\centering
\includegraphics[width=7in]{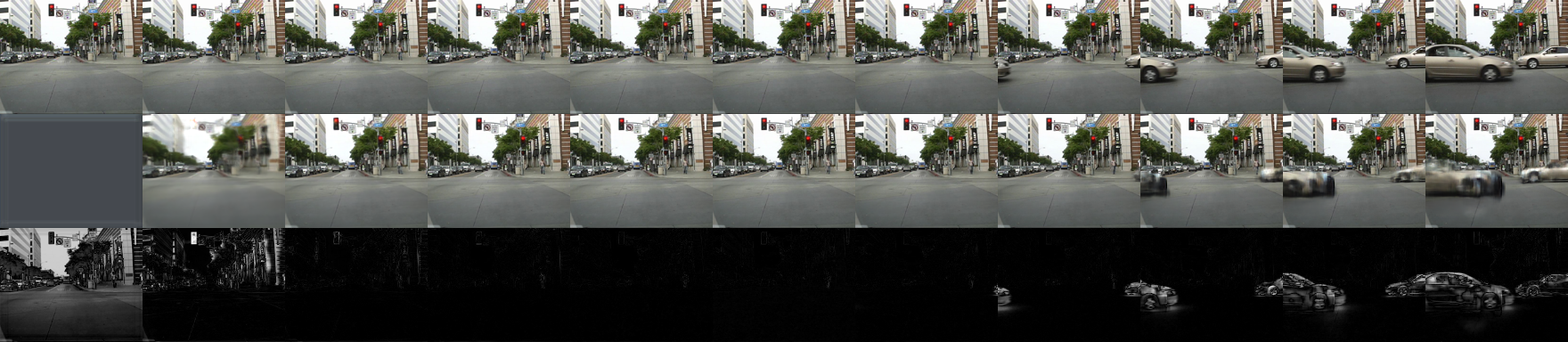}
\end{figure}

\begin{figure}[H]
\centering
\includegraphics[width=7in]{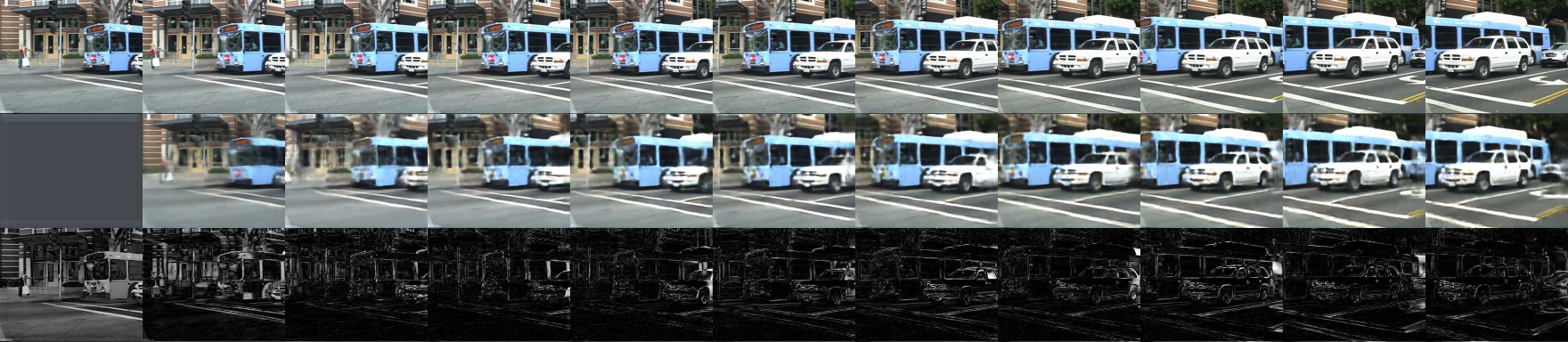}
\end{figure}

\begin{figure}[H]
\centering
\includegraphics[width=7in]{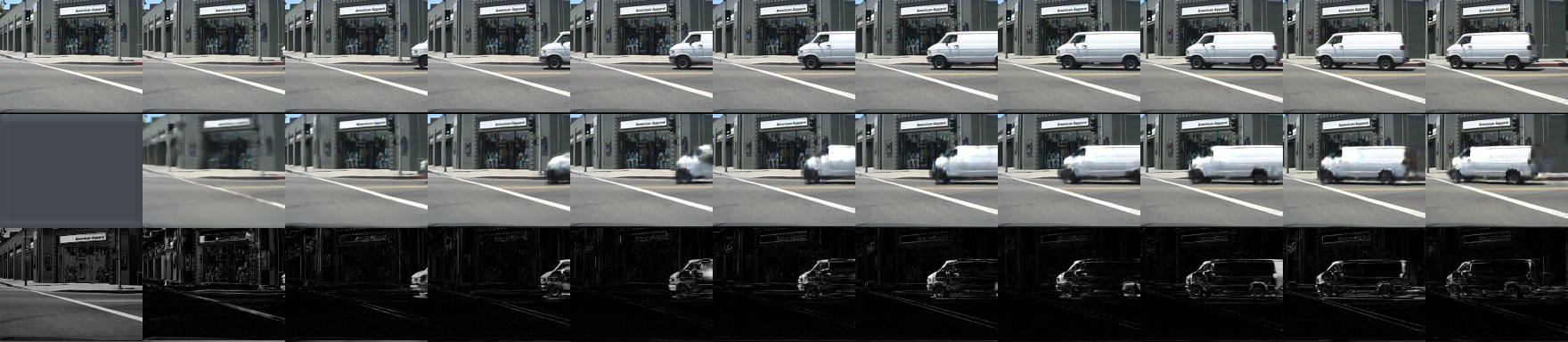}
\end{figure}

\subsection{PreCNet trained on BDD100K (2M)}

\begin{figure}[H]
\centering
\includegraphics[width=7in]{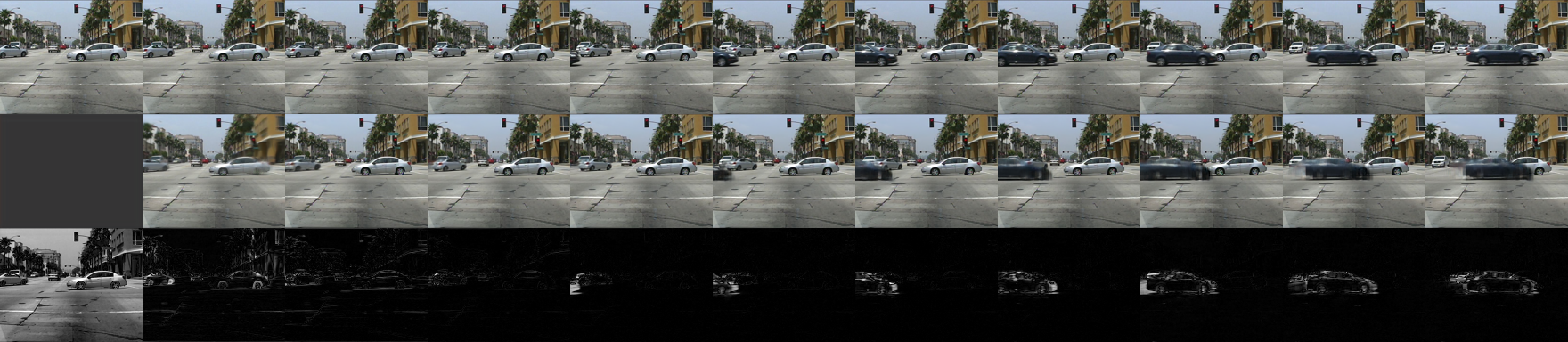}
\end{figure}

\begin{figure}[H]
\centering
\includegraphics[width=7in]{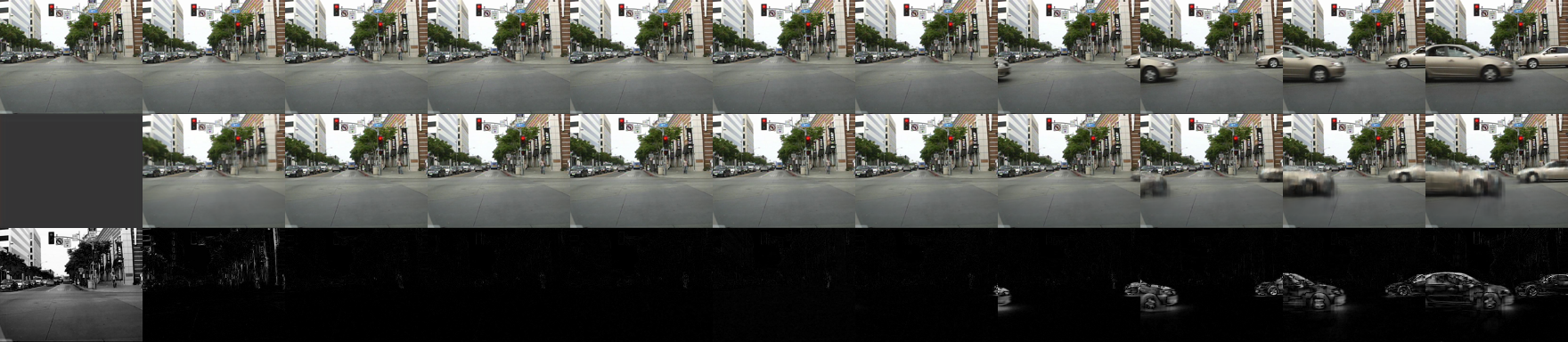}
\end{figure}

\begin{figure}[H]
\centering
\includegraphics[width=7in]{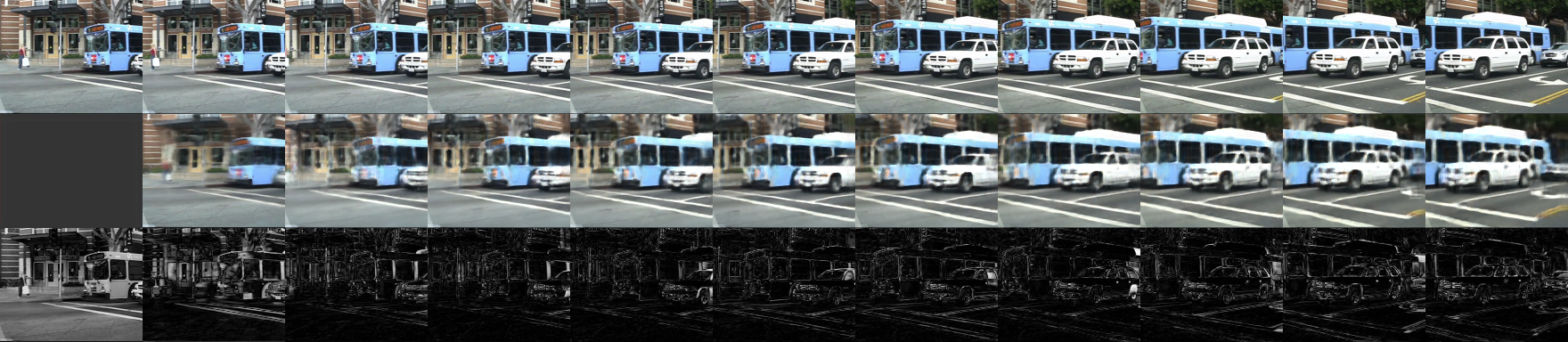}
\end{figure}

\begin{figure}[H]
\centering
\includegraphics[width=7in]{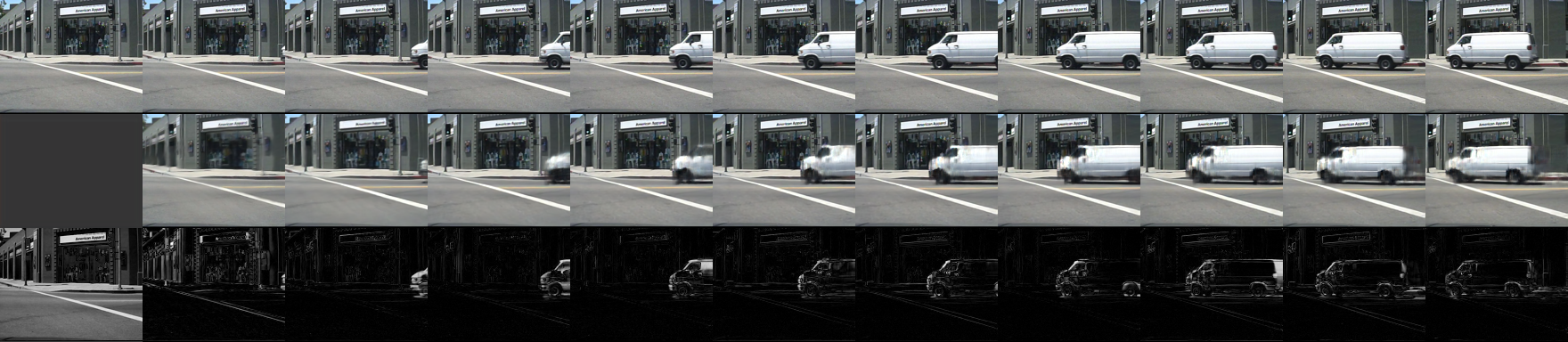}
\end{figure}

\subsection{PreCNet trained on BDD100K (40K)}

\begin{figure}[H]
\centering
\includegraphics[width=7in]{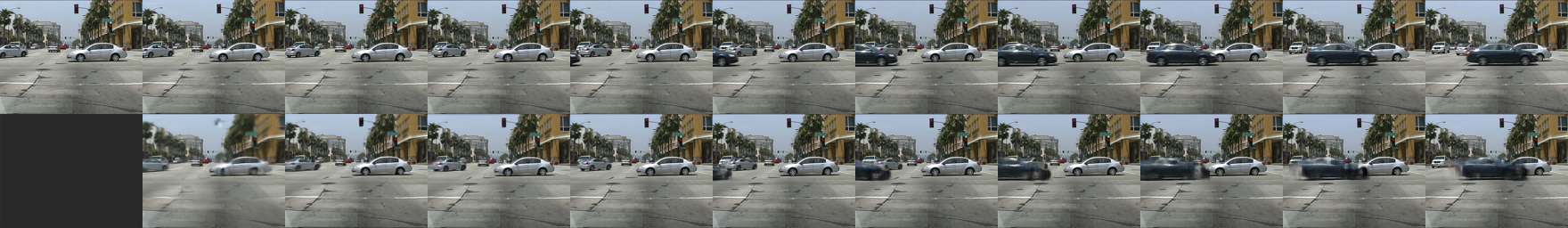}
\end{figure}

\begin{figure}[H]
\centering
\includegraphics[width=7in]{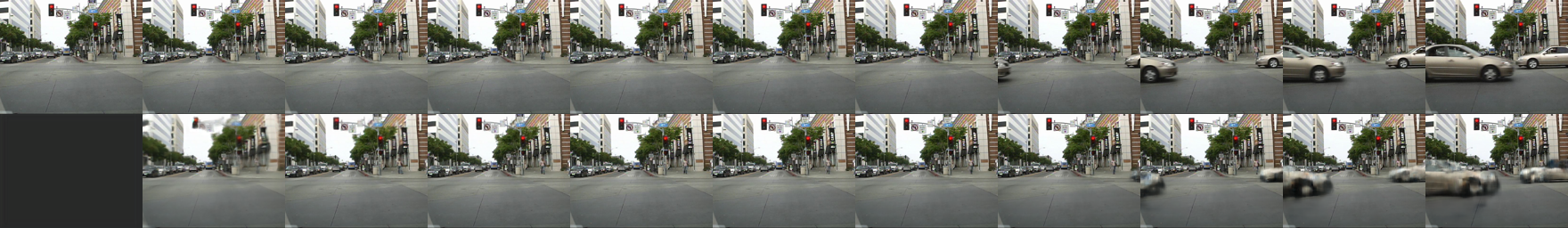}
\end{figure}

\begin{figure}[H]
\centering
\includegraphics[width=7in]{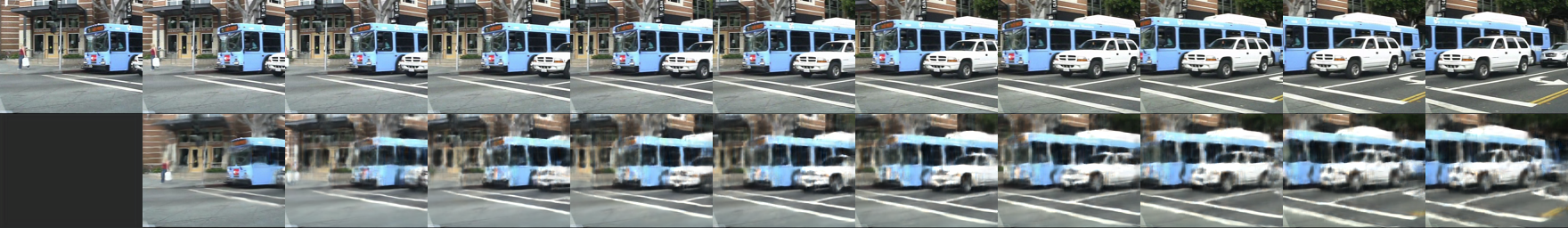}
\end{figure}

\begin{figure}[H]
\centering
\includegraphics[width=7in]{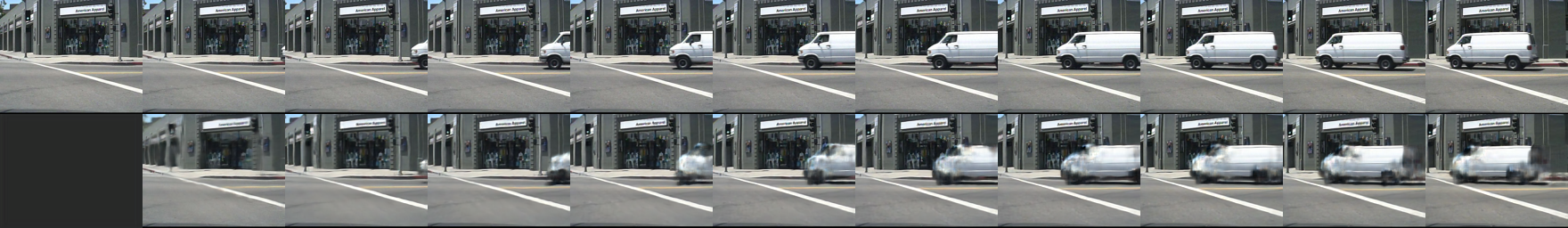}
\end{figure}

\newpage

\section{Multiple frame video prediction sequence}
This figure is analogous to Fig.~8 from the main article, but illustrates a different sequence of images. Location of the sequence in Caltech Pedestrian Dataset: set10-v008.
\begin{figure}[H]
\centering
\includegraphics[width=7in]{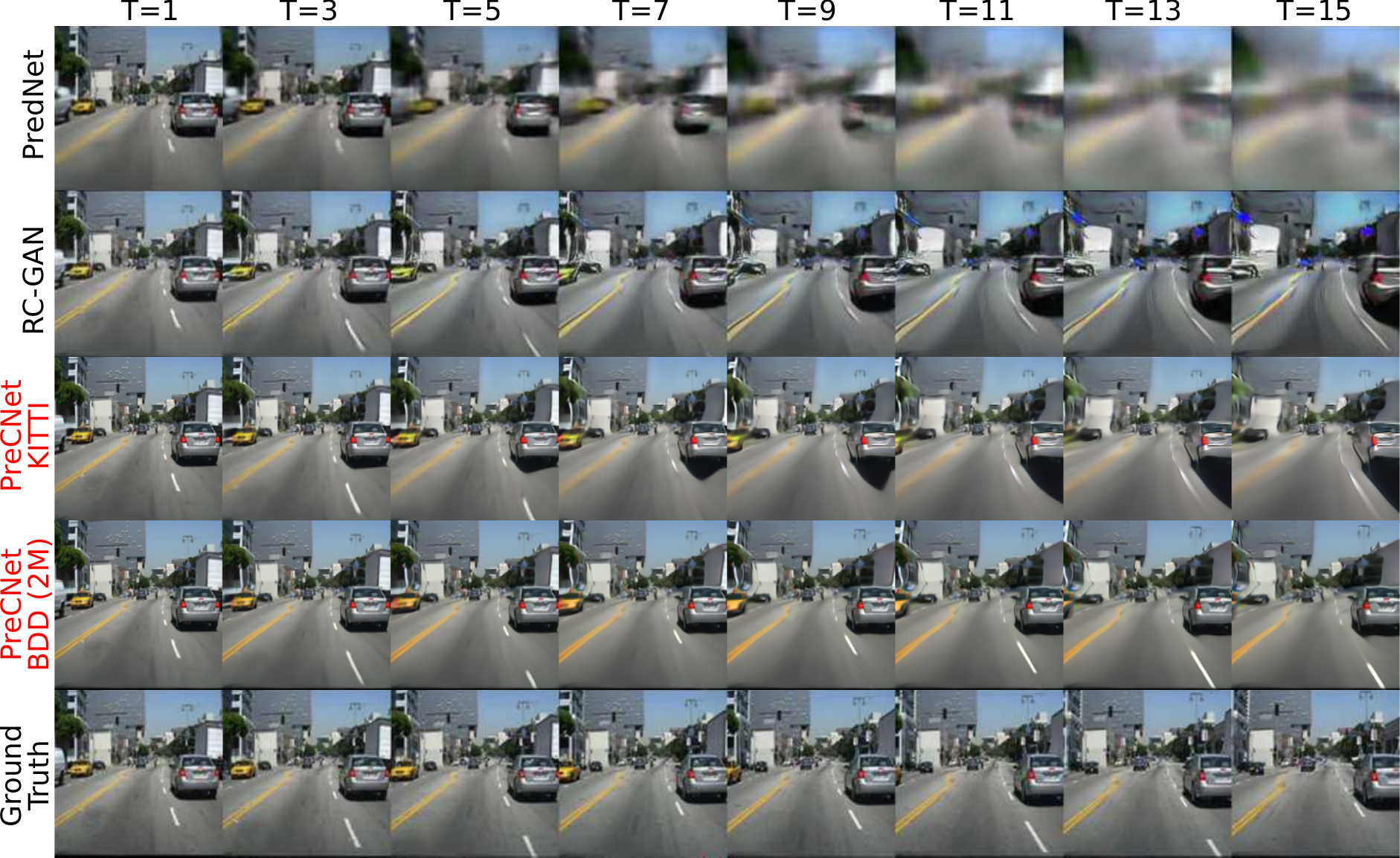}
\end{figure}

\bibliographystyle{IEEEtran}
\bibliography{IEEEabrv,ref}